\documentclass[10pt, journal]{IEEEtran}

\usepackage{xcolor}
\usepackage{amsmath}
\usepackage{amssymb}
\usepackage{graphicx}
\usepackage{booktabs}
\usepackage{siunitx}
\usepackage{standalone}
\usepackage[ruled,vlined]{algorithm2e}
\usepackage{mdframed}
\usepackage{threeparttable}
\usepackage{fancyvrb}
\usepackage{soul}
\usepackage{dsfont,mathabx}
\usepackage{floatrow}
\usepackage{adjustbox}

\begin{document}


\title{Geometric Task Networks: Learning Efficient and Explainable Skill Coordination for Object Manipulation}

\author{Meng Guo and Mathias B\"urger
\thanks{The authors are with Bosch Center for Artificial Intelligence (BCAI), Germany. Corresponding author: Meng Guo. \texttt{Meng.Guo2@de.bosch.com}.}}

\maketitle

\begin{abstract}
Complex manipulation tasks can contain various execution branches of primitive skills in sequence or in parallel under different scenarios.
Manual specifications of such branching conditions and associated skill parameters are not only error-prone due to corner cases but also quickly untraceable given a large number of objects and skills.
On the other hand, learning from demonstration has increasingly shown to be an intuitive and effective way to program such skills for industrial robots.
Parameterized skill representations allow generalization over new scenarios, which however makes the planning process much slower thus unsuitable for online applications.  
In this work, we propose a hierarchical and compositional planning framework that learns a Geometric Task Network (GTN) from exhaustive planners, without any manual inputs.  
A GTN is a goal-dependent task graph that encapsulates both the transition relations among skill representations and the geometric constraints underlying these transitions. 
This framework has shown to improve dramatically the offline learning efficiency, the online performance and the transparency of decision process, by leveraging the task-parameterized models.
We demonstrate the approach on a 7-DoF robot arm both in simulation and on hardware solving various manipulation tasks.
\end{abstract}



\begin{figure*}[t!]
    \centering
    \includegraphics[height=0.3\linewidth]{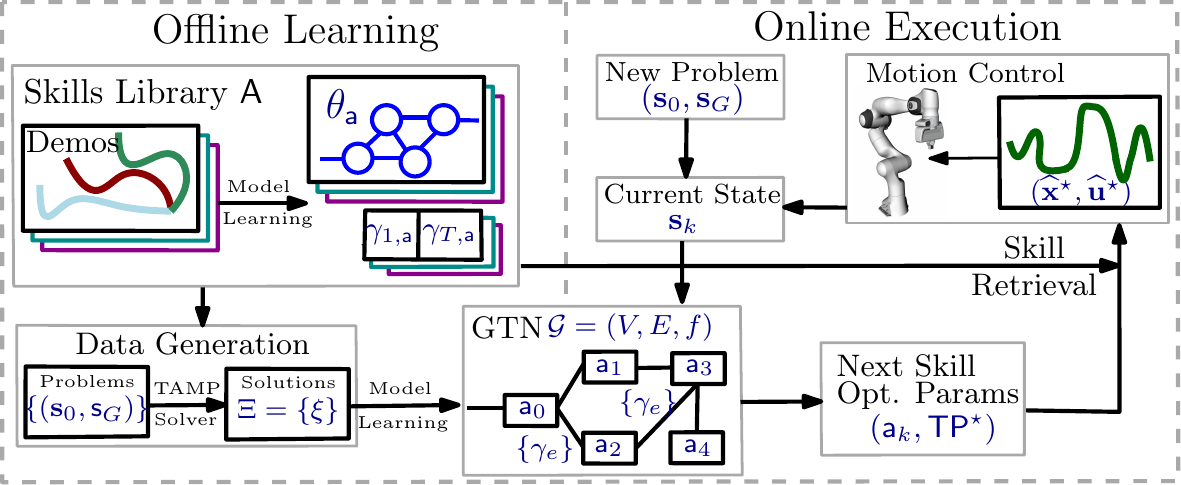}
    \caption{{Illustration of the proposed framework, as described in Sec.~\ref{sec:solution}.
The proposed GTN is first learned offline given demonstrations of primitive skills and a TAMP solver. 
The learned GTN is used online to solve new problems by predicting the next skill and the optimal parameters, given real-time observations.}}
    \label{fig:framework}
    \vspace{-0.15cm}
\end{figure*}

\section{Introduction}\label{sec:intro}

\IEEEPARstart{T}{raditional} manufacturing and industrial practices commonly consist of an array of robotic manipulators, each of which repeats exactly the same trajectory at extremely high speed.
This high efficiency or throughput relies on the fixed environment setup, i.e., every object should be located at the precise pose and any derivation from the setup would lead to failure. 
However, recent years have seen an increasing number of applications that require both flexibility to different setups and generalization to different tasks~\cite{kehoe2015survey}. 
For instance, goods in logistic systems often arrive within unsorted bins; service robots need to deal with unstructured or unseen scenes; products on the assembly line can vary in poses. 
Moreover, instead of a single motion, complex manipulation tasks that contain multiple branches of numerous intermediate skills or sub-tasks should be done by one robot.
As part of the planning process under different scenarios, the suitable branch should be first chosen and then executed with different parameters.
Such adaptation is essential for flexible robotic systems across various applications.
In fact, both self-adaptation and autonomous decision making are important design principles of Industry 4.0 systems from~\cite{hermann2016design}.
On the other hand, learning from demonstration (LfD) has increasingly shown to be an intuitive and effective way to program industrial robots, see~\cite{calinon2016tutorial, Osa2018Imitation}. 
Instead of simply re-playing the recorded demonstration, 
several parameterized skill models have been proposed to improve the generalization over different scenarios.
Such parameterization however makes the above-mentioned planning process harder, due to the high-dimensional parameter space and a large number of skills. 
This consequently leads to long planning time thus unsuitable for online applications. 

The most common approach to tackle the above complexity is by:
(I) limiting the possible scenarios, e.g., by fixing the object poses, such that fewer or only one branch is plausible;
(II) manual definition of branching conditions and parameter selection, e.g., the subspace where one skill is preferred with certain parameters.
However, the first measure would often limit the flexibility and increase the operational cost, 
while the second is prone to engineering errors when covering all corner cases.
Recent works address such problems by task and motion planning (TAMP) that searches through both discrete and continuous space for a path from the initial state and the goal state,
see~\cite{kaelbling2010hierarchical, srivastava2014combined, Toussaint18diff}.
Due to its exponential complexity to the number of skills, the length of tasks, and the dimensionality of parameter space,
the classic TAMP is applicable only to limited industrial applications with ad-hoc heuristics.
Due to the repetitive nature of industrial processes, an intriguing question to ask is ``\emph{Can we solve the problem faster after solving it 100 times?}''.
{Thus some recent work proposes to learn such heuristics from expert solutions or via reinforcement learning, see e.g.,~\cite{spies2019bounded, niekum2015learning, konidaris2012robot}.
However, most of these approaches require enormous amount of training data even for relatively simple tasks.}
This work overcomes the above difficulties by an efficient TAMP framework that is tailored for parametrized skill representations learned from demonstration, without any manual inputs.

In particular, we propose here a hierarchical and compositional planning framework that learns first offline a Geometric Task Network (GTN) from exhaustive planners, which is then used during online execution, as shown in Fig.~\ref{fig:framework}.
This network encapsulates both the transition relations among primitive skill representations and the geometric constraints underlying these transitions, all of which are \emph{parameterized} over the planning goal.
Furthermore, we show that such network can be efficiently combined with skills that are learned via the LfD framework, without any additional manual supervision. 
Notably, both the GTN and the primitive skills are represented based on the task-parameterized  Gaussian mixture models (TP-GMMs) and hidden semi-Markov models (TP-HSMMs). 
This unique combination of homogeneous representations improves dramatically the data efficiency during offline learning, 
reduces significantly the planning time, and ensures the real-time adaptability and failure recovery during online execution.
We validate this approach on a 7-DoF robot arm solving various tasks in simulation and on actual hardware. 

The main contribution of this work is threefold:
(I) it extends the existing work on skill graphs to GTNs, which operate exclusively on skills learned from demonstrations as TP-HSMMs; 
(II) it proposes a novel model for the constraints embedded in the GTNs, based on TP-GMMs.
By exploiting the geometric structure of demonstrated skills w.r.t. relevant object frames, the GTNs can be learned and executed in an extremely time-and-data-efficient way, without the necessity of a simulator;
(III) this task-parametrized model provides a fully-explainable planner regarding why a skill and these parameters are chosen, what the alternatives are, and how confident such options are, 
thus yielding better transparency with the human operators during the decision process.


\begin{figure*}[t!]
\centering
\includegraphics[height=0.24\linewidth]{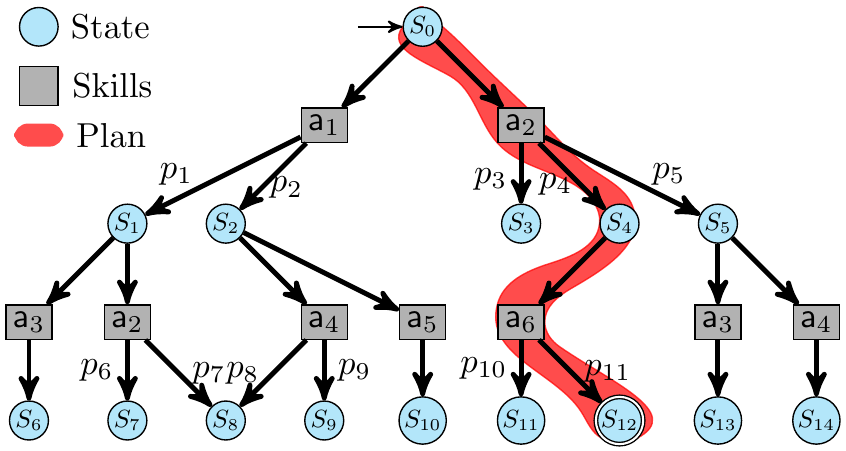}
\hspace{0.1in} \vrule \hspace{0.1in}
\includegraphics[height=0.24\linewidth]{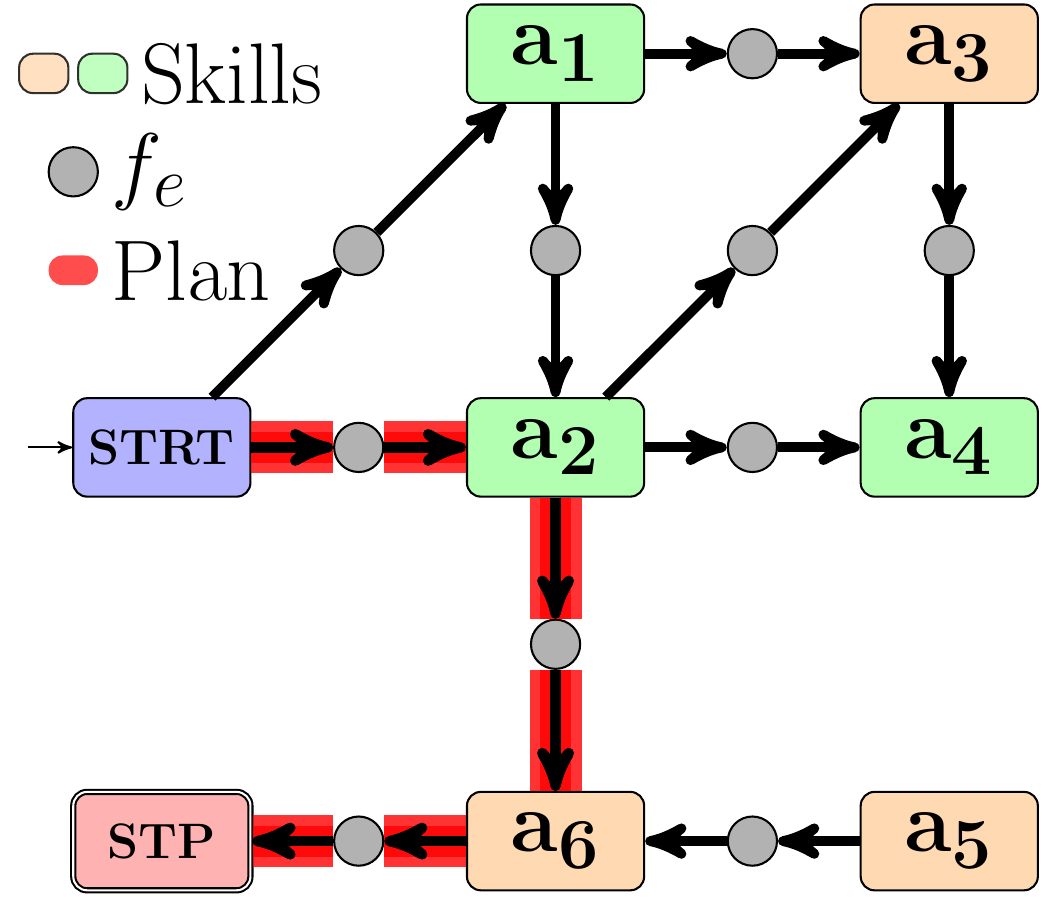}
\caption{
Comparison of structure between the classic state graph (\textbf{left}) and the proposed geometric task network GTN (\textbf{right}), as discussed in Sec.~\ref{sec:tamp-review}. 
Note that the corresponding plans are highlighted in red in both structures.}
\label{fig:search}
\end{figure*}

\section{Related Work}\label{sec:related}

\subsection{Learning from Demonstration (LfD)}\label{sec:lfd-review}
Compared with traditional motion planning from~\cite{lavalle2006planning}, LfD is an intuitive way to transfer human skills to robots, see~\cite{calinon2016tutorial, Osa2018Imitation, huang2019neural}.
Various teaching methods can be used such as kinesthetic teaching in~\cite{calinon2016tutorial}, tele-operation in~\cite{abbeel2004apprenticeship}, and visual demonstration in~\cite{huang2019neural}. 
Different skill models are proposed to abstract these demonstrations: 
full trajectory of robot end-effector in~\cite{Osa2018Imitation}, 
dynamic movement primitives (DMPs) in~\cite{rozo2016learning},
TP-GMMs in~\cite{calinon2016tutorial, Zeestraten17riemannian} which extend GMMs by incorporating observations from different perspectives (so called task parameters),
TP-HSMMs in~\cite{Schwenkel2019Optimizing, rozo2020learning},
and deep neural networks~\cite{huang2019neural} that map directly observations to control inputs.
In this work, we adopt the representation as TP-HSMMs, mainly due to two reasons:
first TP-HSMMs provide an elegant probabilistic representation of motion skills, which extracts both temporal and spatial features from few human teachings. 
In contrast, DMPs and TP-GMMs can only encode spatial information;
second, task parameterization allows the model to generalize to new situations, see e.g., applications in collaborative transportation~\cite{rozo2016learning}.
This work further extends this representation with \emph{grounded} precondition and effect models, which provides symbolic relations among the skills, but without direct symbol generation and logic reasoning.

\subsection{Task and Motion Planning}\label{sec:tamp-review}
Task planning focuses on constructing a discrete high-level plan via abstract decision-making, e.g., via logic-reasoning from~\cite{fox2003pddl2}. 
On the other hand, motion planning addresses the low-level sensing and control problem of a dynamic system, e.g., as reviewed in~\cite{lavalle2006planning}.
The area of task and motion planning (TAMP) attempts to improve the synergies between them.
As illustrated in Fig.~\ref{fig:search}, the planning process over the state graph consists of searching over both {the set of discrete} skill primitives and the continuous skill parameters. 
One direct challenge that arises is that state abstractions along with geometric constraints are difficult to capture symbolically and more so automatically, as also mentioned in~\cite{srivastava2014combined}.
{A common approach is to combine discrete logic reasoning with sampling-based motion planners, e.g., in~\cite{srivastava2014combined, Toussaint18diff, wang2021learning}.
Despite the intuitive planning process and direct interface with existing motion planners, 
such approaches often suffer from the intensive manual modeling effort and the exponential search space.}
Different heuristics are proposed to prune the search tree thus reduce the motion planning complexity, see~\cite{Toussaint18diff, konidaris2018skills}.

A conjugate view of the state graph is the so-called skill graph as shown in Fig.~\ref{fig:search}, where instead the nodes are primitive skills and edges are implicit state abstractions, see~\cite{niekum2015learning, konidaris2012robot, huang2019neural,  hayes2016autonomously,  frazzoli2005maneuver}. 
Skill graphs are also used for the purpose of TAMP in these work.  
In contrast, an optimal plan in the skill graph can have cycles and repetitive segments, i.e., not necessarily the shortest path from start node to stop node. 
The work in \cite{hayes2016autonomously} extends the hierarchical task networks (HTN) to conjugate task graph (CTG) without any parameterization on the skill primitives.   
{Similar idea is used in \cite{coles2010forward} for partial-order planning.}
Moreover, \cite{frazzoli2005maneuver} calls such graph as maneuver automaton, which however is \emph{manually} designed instead of learned, whereas~\cite{huang2019neural} require similar structural supervision during training.
The works in~\cite{niekum2015learning, konidaris2012robot} learn such task-level graph from complete demonstrations of the \emph{whole} manipulation task and with manual specification of action sequences during training. 
Similarly, the work in~\cite{huang2019neural} learns a flexible network based on Neural Programmer Interpreter from~\cite{reed16neural} to allow generalization over new tasks. 
The method in~\cite{konidaris2012robot} relies on ``change point'' detection to segment these task demonstrations with simple non-parameterized models in 2D,
while~\cite{niekum2015learning} assumes each skill primitive is parameterized to only \emph{one} object frame.
In this work, we adopt this conjugate perspective but only require demonstrations of primitive skills \emph{locally} without any specific task in mind. 
The main advantage is that such skills can be shared and re-used across different tasks.
Furthermore, compared with~\cite{niekum2015learning, konidaris2012robot}, 
the skill representation used here as TP-HSMMs is more \emph{general} since it incorporates temporal properties within the demonstration and 
allows an arbitrary number of relevant frames. 
Lastly, the constraints embedded in the proposed GTN is based on TP-GMMs,  which are extremely efficient for learning and inferring when combined with the TP-HSMMs of the skill model.
To the best of the authors' knowledge, this feature has not been exploited in literature.

\begin{figure*}[th!]
\centering
\includegraphics[width=0.8\linewidth]{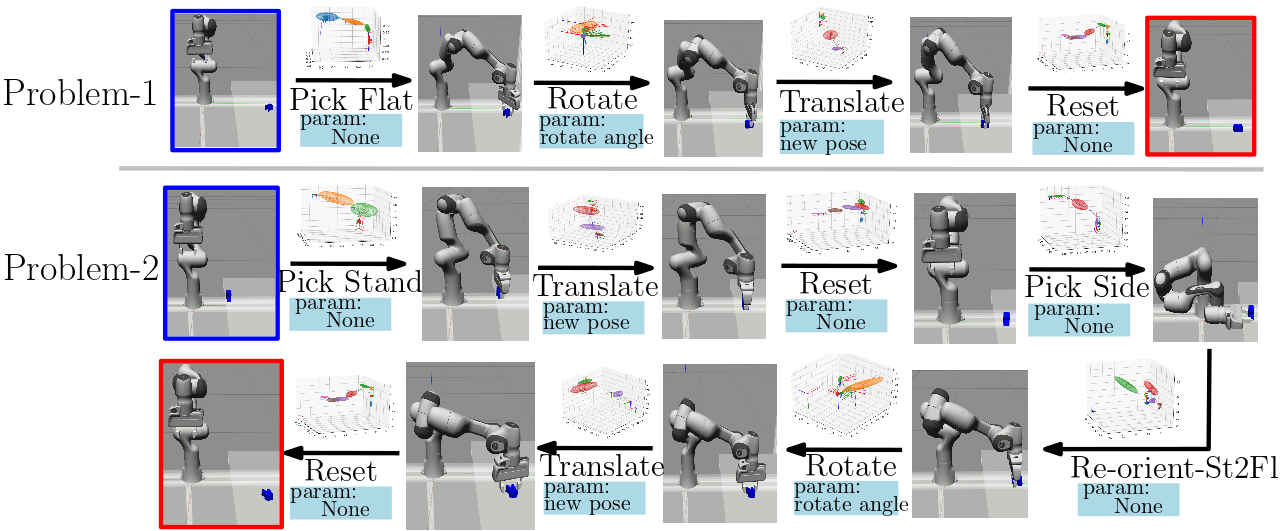}
\caption{
Examples of solutions to two problems of the \emph{same} task, where goal states are highlighted in red. 
Note that snapshots represent the intermediate states, where the transitions are driven by the chosen skill under desired parameters.
The associated skill model and reference trajectories are also shown.}
\label{fig:tamp-solutions}
\end{figure*}

\subsection{Imitation Learning}\label{sec:hil-review}
An important technique used in this work is to learn an imitation policy from an expert solver or human, such that this policy can be used online but with less solution time or improved generalization.
Imitation learning has been widely used for various purposes, e.g., autonomous driving, 
robot motion control~\cite{huang2019neural, kim2019learning}, and multi-robot coordination~\cite{spies2019bounded, tamar2016value}.
Most of the these work has a strong focus on learning low-level control policy from raw visual inputs, without considering high-level tasks.
Furthermore, training data can be generated from a complete solver~\cite{spies2019bounded, kim2019learning, kroemer2015towards} or expert demonstrations~\cite{konidaris2012robot}.
They are commonly represented by deep neural networks (DNN) such as CNN~\cite{spies2019bounded, tamar2016value}, GNN~\cite{kim2019learning}, and VAE~\cite{lynch2019play}.
High-dimensional sensory inputs such as images in~\cite{huang2019neural, tamar2016value} or point clouds in~\cite{konidaris2018skills} enable direct reasoning over raw inputs, while direct state information such as object poses in~\cite{kim2019learning} provides easy interfaces to the existing motion planners.
Most of the work above requires large amount of training data for the \emph{complete} task, mainly due to the training mechanism of DNN and the high-dimensional state space. 
In contrast, the geometric relation among the robot and the objects in the proposed GTNs are modeled using TP-GMMs, which are more efficient to learn.

Furthermore, generalization to new scenarios or even new tasks is an important aspect of imitation learning. 
{Namely, the learned policy should not simply replicate what was taught during the training, rather to abstract the underlying reasoning and methods. 
Such generalization can happen at different levels, e.g., }
(I) at the skill-level that the execution of single primitive skill can adapt to different scenarios such as different robot and object poses, e.g., via end-to-end vision-guided control~\cite{huang2019neural}, motion constraints for sampling planners~\cite{kim2019learning}, task parameterization~\cite{Schwenkel2019Optimizing, rozo2020learning} and manual classification~\cite{niekum2015learning}.
(II) at the task-level that different instances of the same task, or even different tasks can be solved. 
For instance,~\cite{huang2019neural} considers rather homogeneous tasks with uniform sequences of a small number of primitive skills, such as sorting with ``pick-and-place'', where the generalization is w.r.t. the number of objects;
task generalization in~\cite{hayes2016autonomously} relies heavily on the initial task network which is manually  defined.
In this work, we address mainly complex manipulation tasks where the sequence of desired skills and their parameters changes significantly under different pairs of initial and goal states. 

\section{Preliminaries on Task-Parameterized Models}\label{sec:preliminary}

This section presents briefly the essential background on task-parameterized models. 
More details are given in the supplementary file and see~\cite{calinon2016tutorial, Zeestraten17riemannian,  Schwenkel2019Optimizing, rozo2020learning}.

\subsection{TP-GMMs}\label{subsec:tp-gmm}
A task parameterized Gaussian mixture model (TP-GMM) is described by the following model parameters 
$\boldsymbol{\gamma} = \{\pi_k,\{\boldsymbol{\mu}_k^{(p)},\boldsymbol{\Sigma}_k^{(p)}\}_{p=1}^P\}_{k=1}^K$,
where $K$ represents the number of Gaussian components in the mixture model, 
$\pi_k$ is the prior probability of each component, 
and $\{\boldsymbol{\mu}_k^{(p)},\boldsymbol{\Sigma}_k^{(p)}\}_{p=1}^P$ are the mean and covariance of the $k$-th Gaussian component within frame~$p$. 
Frames provide observations of the \emph{same} data but from different perspectives, the instantiation of which is called a {task parameter}.
Differently from standard GMM used in~\cite{niekum2015learning}, the mixture model above can not be learned independently for each frame. 
Given these observations, the Expectation-Maximization (EM) method from~\cite{dempster77em} is a well-established method to learn such models with polynomial complexity to the state dimension and the number of components.

Afterwards, given a set of new frames $\{\boldsymbol{b}^{(p)},\boldsymbol{A}^{(p)}\}_{p=1}^P$, the learned TP-GMM is converted into one single GMM with parameters $\{\pi_k, (\boldsymbol{\hat{\mu}}_{k}, \boldsymbol{\hat{\Sigma}}_{k})\}_{k=1}^K$, by multiplying the affine-transformed Gaussian components across different frames.
{Note that \emph{obstacles} can be explicitly added as frames in the above model, 
or implicitly considered in the demonstrations.}

\subsection{Complete Skill Model}\label{subsec:tp-hsmm}
As proposed in our earlier work~\cite{Schwenkel2019Optimizing},  the complete skill model of~$\mathsf{a}$ can be learned:
\begin{equation}\label{eq:skill-model}
\mathcal{M}_{\mathsf{a}} = (\boldsymbol{\theta}_\mathsf{a},\, \boldsymbol{\boldsymbol{\gamma}}_{1, \mathsf{a}}, \, \boldsymbol{\boldsymbol{\gamma}}_{T,\mathsf{a}}),
\end{equation}
where the elements are as follows.

(I) The \emph{trajectory} model~$\boldsymbol{\theta}_{\mathsf{a}}$ 
as the task-parameterized Hidden semi-Markov Model (TP-HSMM) that encapsulates both temporal and spatial property of all human demonstrations~$\mathsf{D}_{\mathsf{a}}$. 
A TP-HSMM  is built upon a TP-GMM from Sec.~\ref{subsec:tp-gmm} above. 
HSMM extends the standard HMM by embedding temporal information of the underlying stochastic process. 
This means that a transition to the next component depends on the current component as well as on the elapsed time since it was entered.
More specifically, a TP-HSMM is defined as:
\begin{equation}\label{eq:tp-hsmm}
\boldsymbol{\theta} = \left\{ \{a_{hk}\}_{h=1}^K,\, (\mu_k^D, \sigma_k^D),\, \boldsymbol{\gamma}_k \right\}_{k=1}^K,
\end{equation}
where $a_{hk}$ is the transition probability from component~$h$ to~$k$; $(\mu_k^D, \sigma_k^D)$ describe the Gaussian duration of component~$k$; and $\boldsymbol{\gamma}_k$ is component~$k$ as a TP-GMM above.

(II) The \emph{precondition} model~$\boldsymbol{\boldsymbol{\gamma}}_{1, \mathsf{a}}$
and the \emph{effect} model~$\boldsymbol{\boldsymbol{\gamma}}_{T,\mathsf{a}}$. 
Note that $\boldsymbol{\boldsymbol{\gamma}}_{1, \mathsf{a}}$ are the TP-GMMs that model the system state \emph{before} executing the skill; 
$\boldsymbol{\boldsymbol{\gamma}}_{T,\mathsf{a}}$ are the TP-GMMs that model the change of system state \emph{after} executing the skill. 
{They are used to model the geometric relations among the robot and all objects, e.g., 
where the objects are w.r.t. each other and the end-effector. 
Note that they are \emph{not} the first and last component of the TP-HSMM model $\boldsymbol{\theta}$ above. }
These models are automatically \emph{grounded} to system state,
without the need for manual definitions as in~\cite{srivastava2014combined, Toussaint18diff}.

All three elements in~\eqref{eq:skill-model} are task parameterized. 
These task parameters can be either determined by the system state (such as object poses) or \emph{freely} chosen.
While the first case are commonly seen in literature, the latter case is often tackled by hand-crafted rules. 
However, such free task parameters are essential for the flexible skills such as translation and rotation.
Detailed derivations are given in~\cite{Schwenkel2019Optimizing}, also summarized in the supplementary file.

\begin{figure*}[th!]
\centering
\includegraphics[width=0.95\linewidth]{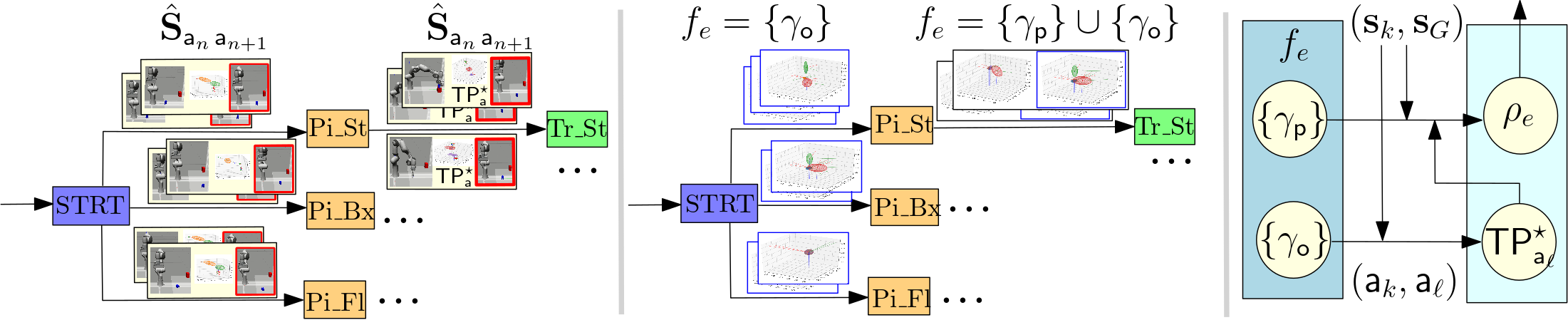}
\caption{
The learning process of GTN from the gathered data, as described in Sec.~\ref{subsubsec:learn}.
\textbf{Left}: preliminary structure of GTN where edges are labeled by the augmented states~$\widehat{\mathbf{s}}_{\mathsf{a}_{n}\mathsf{a}_{n+1}}$, 
where goal states are highlighted in red and the executed trajectory is plotted along with the learned TP-HSMMs. 
\textbf{Middle}: the learned GTN, where edges are labeled by the geometric constraints~$f_e$ as a set of TP-GMMs.
Note that some skills have no free task parameters and only $\{\boldsymbol{\gamma}_{\mathsf{o}}\}$ is learned (in yellow), while~$\{\boldsymbol{\gamma}_{\mathsf{p}}\}$ is learned in addition for skills with free parameters (in green). 
\textbf{Right}: how the learned $f_e$ is used to compute the score $\rho_e$ for each outgoing transition, given the current state $\mathbf{s}_k$ and goal state $\mathbf{s}_G$.
}
\label{fig:learn-gtn}
\end{figure*}

\section{Problem Description}\label{sec:problem}

Consider a multi-DoF robotic arm, of which the end-effector has state $\boldsymbol{r}$ such as its 6-D pose and gripper state.
We assume that the robot operates within a static and known workspace.
Also, within the reach of the robot, there are multiple objects of interest denoted by $\mathsf{O}=\{\mathsf{o}_1, \mathsf{o}_2,\cdots,\mathsf{o}_J\}$. 
Each object is described by its state~$\boldsymbol{p}_{\mathsf{o}}$ such as its 6-D pose.

Moreover, there is a set of \emph{primitive} manipulation skills that enable the robot to manipulate these objects, denoted by $\mathsf{A}=\{\mathsf{a}_1,\mathsf{a}_2,\cdots,\mathsf{a}_H\}$.  
For each skill, a human user performs several kinesthetic  demonstrations on the robot.
Particularly, for skill $\mathsf{a}\in \mathsf{A}$, the set of objects involved is given by $\mathsf{O}_{\mathsf{a}} \subseteq \mathsf{O}$ and the set of demonstrations is given by $\mathsf{D}_{\mathsf{a}}=\{\mathsf{D}_1,\cdots, \mathsf{D}_{M_{\mathsf{a}}}\}$, where each demonstration $\mathsf{D}_m$ is a \emph{timed} sequence of states $\mathbf{s}\in \mathbf{S}$ that consists of the end-effector state $\boldsymbol{r}$ and object states $\{\boldsymbol{p}_{\mathsf{o}}, \, \mathsf{o}\in \mathsf{O}_{\mathsf{a}}\}$, i.e.,
$\mathsf{D}_m = \big[\mathbf{s}_t\big]_{t=1}^{T_m} = \left[\big(\boldsymbol{r}_t, \,\{\boldsymbol{p}_{t,\mathsf{o}}, \, \mathsf{o}\in \mathsf{O}_{\mathsf{a}}\} \big)\right]_{t=1}^{T_m}$.
Via a combination of these skills, the objects can be manipulated to reach different states. 

We consider a generic manipulation \emph{task}, which however consists of many \emph{problem instances}. 
Each problem instance is specified by an initial state~$\mathbf{s}_0$ and a set of desired goal states $\{\mathbf{s}_G\}$. 
A problem is solved when the robot successfully modified the system state from $\mathbf{s}_0$ to $\{\mathbf{s}_G\}$.
{Two problems belong to the same task if they can be solved by the same set of primitive skills as introduced above}.
We are interested in solving complex manipulation tasks where the sequence of desired skills and their parameters changes significantly for different problems. 
An example of solutions to different problem instances is shown in Fig.~\ref{fig:tamp-solutions}.

Thus, our objective is to learn a GTN as shown in Fig.~\ref{fig:search} for the considered task that: 
given any feasible problem instance of this task, 
the learned GTN generates online:
(I) the sequence of skills with the associated parameters;
and (II) the associated reference trajectory for each skill within the sequence,
such that the system reaches the goal states $\{\mathbf{s}_G\}$ from~$\mathbf{s}_0$.


\section{Proposed Solution}\label{sec:solution}
Main components of the proposed framework as depicted in Fig.~\ref{fig:framework} are presented in this section.
The offline learning of GTNs are described in Sec.~\ref{subsec:learn-GTN}, 
while the online execution of learned GTNs given new tasks is described in Sec.~\ref{subsec:execution-GTN}.

\subsection{Offline Learning}\label{subsec:learn-GTN}
This section describes first the general structure of GTNs, 
then the procedure to gather training data, and lastly the algorithms to learn the associated GTN for a certain task.

\subsubsection{Structure of GTNs}\label{subsubsec:structure-GTN}
As illustrated in Fig.~\ref{fig:search}, a GTN has a relatively simple structure defined by the triple $\mathcal{G}=(V,\, E,\, f)$.
The set of nodes~$V$ is a subset of the primitive skills $\mathsf{A}$;
the set of edges $E \subseteq V \times V$ contains the allowed transitions from one skill to another;
the function $f: E \rightarrow 2^{\boldsymbol{\boldsymbol{\gamma}}_{\mathbf{s}}}$ maps each edge to a \emph{set} of TP-GMMs over the system state $\mathbf{s}$.
The structure of TP-GMMs $\boldsymbol{\boldsymbol{\gamma}}$ is defined in Sec.~\ref{sec:preliminary}.
Intuitively, $(V,\, E)$ specifies how skills can be executed consecutively for the given task,
while function $f_e = f(e)$ models the geometric constraints among the objects and the robot for each edge $e\in E$, which is explicitly conditioned on the goal state of the problem. 
The detailed representation of $f(\cdot)$ is given in the sequel.


\subsubsection{Training Data Gathering}\label{subsubsec:train-data}
First, for each primitive skill~$\mathsf{a}$, the skill model $\mathcal{M}_{\mathsf{a}}$ from~\eqref{eq:skill-model} has to be learned. 
Namely, the trajectory model~$\boldsymbol{\theta}_\mathsf{a}$ is used to retrieve the trajectory, while the precondition and effect models $\boldsymbol{\boldsymbol{\gamma}}_{1, \mathsf{a}}, \, \boldsymbol{\boldsymbol{\gamma}}_{T,\mathsf{a}}$ are used to compute the initial condition, and predict the resulting state. 

Secondly, a {complete} TAMP over the state space is used to generate the training data for learning GTNs.
As shown in Fig.~\ref{fig:search}, 
starting from the initial state~$\mathbf{s}_0$, a graph search algorithm is used to traverse the state space until the goal state is reached, e.g., breadth-first search or Dijkstra or $A^{\star}$ algorithm see~\cite{lavalle2006planning}.
{Note that any state-of-the-art TAMP planners can be used here.
In fact, the faster each problem in the training set is solved, 
the less time the data gathering process takes.}

At each visited state~$\mathbf{s}_i$, any skill $\mathsf{a}_k\in \mathsf{A}$ can be executed to drive the system to a new state~$\mathbf{s}_j$.
{The associated parameters (denoted by~$\mathsf{TP}_{\mathsf{a}_k}$) are either computed directly from the current state~$\mathbf{s}_i$ (e.g., object poses), or {sampled} uniformly (e.g., free task parameters) from the precondition model~$\boldsymbol{\boldsymbol{\gamma}}_{1,\mathsf{a}_k}$ in~\eqref{eq:skill-model}.}
Moreover, given these parameters, the resulting state~$\mathbf{s}_j$ is computed as the {mean} of the effect model~$\boldsymbol{\boldsymbol{\gamma}}_{T,\mathsf{a}_k}$.
Note that the same skill with {different} parameters drives the system to different states.
In this way, both the discrete choices of skills and continuous choices of skill parameters are explored during the search, yielding a \emph{hybrid} method.
Thus each edge is described by the ``state-action-state'' triple~$(\mathbf{s}_i, (\mathsf{a}_k, \mathsf{TP}_{\mathsf{a}_k}), \mathbf{s}_j)$, of which the cost is computed based on the confidence measure from~\cite{Schwenkel2019Optimizing}.
Other choices are also available as shown numerically in Sec.~\ref{sec:experiments}. 
Once one of the goal states in $\{\mathbf{s}_G\}$ is reached, a complete plan is retrieved as:
\begin{equation}\label{eq:complete-plan}
\boldsymbol{\xi} =\mathbf{s}_0 \,(\mathsf{a}_0, \mathsf{TP}_{\mathsf{a}_0})\,\mathbf{s}_1 \,(\mathsf{a}_1, \mathsf{TP}_{\mathsf{a}_1})\mathbf{s}_2 \cdots \mathbf{s}_G, 
\end{equation}
of which  the associated discrete plan is denoted by~$\boldsymbol{\xi}_d = \mathsf{a}_0\mathsf{a}_1\cdots \mathsf{a}_G$. 
The above procedure is repeated for different initial and goal states (i.e., problems) of the same task, of which the resulting plans are stored in a dataset, denoted by $\boldsymbol{\Xi}=\{\boldsymbol{\xi}\}$. 
Despite of being asymptotically complete, the above planner has \emph{exponential} complexity w.r.t. the number of skills, the dimensionality of their parameters and the plan length. 
Last but not least, different from most related work~\cite{ huang2019neural, kim2019learning}, the above planner does \emph{not} rely on a simulator to evaluate the resulting state or the associated cost after executing a skill.

Two examples are shown in Fig.~\ref{fig:tamp-solutions}.
It can be seen that the solutions to two problems of the \emph{same} task can be completely different, e.g., in case (a), the alphabet can be directly picked, rotated and translated to the desired goal, while in case (b), the alphabet should be picked, moved and re-oriented from ``standing'' to ``lying flat''. 
Notice that some skills are executed with different parameters in different solutions, e.g., skill \texttt{rotate} in case (a) and (b), or even in the same solution, e.g., skill \texttt{translate} in case (b).

\subsubsection{GTN Model Learning.}\label{subsubsec:learn}
For each plan $\boldsymbol{\xi} \in \boldsymbol{\Xi}$,
a virtual start action~$\underline{\mathsf{a}}$ is prepend to the beginning of $\boldsymbol{\xi}$ and a virtual end action~$\overline{\mathsf{a}}$ is append to its end, i.e.,
$
\boldsymbol{\xi} = \underline{\mathsf{a}} \mathbf{s}_0 (\mathsf{a}_0, \mathsf{TP}_{\mathsf{a}_0})\mathbf{s}_1 \cdots (\mathsf{a}_{N-1}, \mathsf{TP}_{\mathsf{a}_{N-1}}) \mathbf{s}_G  \overline{\mathsf{a}},
$
where the plan has length $N$ and $\mathbf{s}_G$ is the \emph{final} goal state reached by the plan.
Then, for each ``action-state-action'' triple $\big{(}(\mathsf{a}_n, \mathsf{TP}_{\mathsf{a}_n}), \mathbf{s}_{n+1}, (\mathsf{a}_{n+1}, \mathsf{TP}_{\mathsf{a}_{n+1}})\big{)}$ within $\boldsymbol{\xi}$,
the pair $(\mathsf{a}_{n},  \mathsf{a}_{n+1})$ is added to the edge set $\widehat{E}$ if not already present; and 
a new \emph{augmented} state $\widehat{\mathbf{s}}= (\mathbf{s}_{n+1}, \mathsf{TP}_{\mathsf{a}_{n+1}},\mathbf{s}_G)$ is added to the set  of all augmented states, denoted by~$\widehat{\mathbf{s}}_{\mathsf{a}_{n}\mathsf{a}_{n+1}}$, for each {unique} skill transition $(\mathsf{a}_{n},\mathsf{a}_{n+1})$.

\begin{algorithm}[t]
   \caption{Offline Learning of GTN} \label{alg:offline-learn}
   \LinesNumbered
   \DontPrintSemicolon
   \KwIn{$\mathsf{D}_{\mathsf{a}}, \, \forall \mathsf{a} \in \mathsf{A}$.}
   \KwOut{$\boldsymbol{\Xi}$, $\mathcal{G}=(V, E, f)$.}
   Learn $\mathcal{M}_{\mathsf{a}}$ in~\eqref{eq:skill-model} for each $\mathsf{a} \in \mathsf{A}$.\;
   \ForAll{$(\mathbf{s}_0,\, \mathbf{s}_G)$}{
       Find the complete plan $\boldsymbol{\xi}$ in~\eqref{eq:complete-plan} using an exhaustive TAMP.\;
       Add $\boldsymbol{\xi}$ to $\boldsymbol{\Xi}$.\;
   }
   Construct $\widehat{E}$ and $\{\widehat{\mathbf{s}}_{e}, \, \forall e\in \widehat{E}\}$ given $\boldsymbol{\Xi}$.\;
   Build $(V,\, E)$ given $\widehat{E}$.\;
   Compute $f(e)$ by~\eqref{eq:f} given $\widehat{\mathbf{s}}_e$, $\forall e\in \widehat{E}$.\;
   \vspace{-0.1cm}
\end{algorithm}

Once all plans within $\boldsymbol{\Xi}$ are processed, the GTN $\mathcal{G}$ can be constructed as follows.
First, its vertices and edges are directly derived from $\widehat{E}$.
Then, for each $(\mathsf{a}_k, \mathsf{a}_\ell) \in \widehat{E}$,
the function $f(\mathsf{a}_k, \mathsf{a}_\ell)$ returns a \emph{set} of TP-GMMs that are computed from $\widehat{\mathbf{s}}_{\mathsf{a}_k \mathsf{a}_\ell}=\{(\mathbf{s}_j, \mathsf{TP}_{\mathsf{a}_\ell}, \mathbf{s}_G)\}$, given by:
\begin{equation}\label{eq:f}
f(\mathsf{a}_k, \mathsf{a}_\ell)=\{\boldsymbol{\gamma}_{\mathsf{p}}, \forall \mathsf{p}\in \mathsf{TP}_{\mathsf{a}_\ell}\}\cup \{\boldsymbol{\gamma}_{\mathsf{o}}, \forall \mathsf{o} \in \mathsf{O}_{\mathsf{a}_\ell}\},
\end{equation}
where 
(I) For each task parameter $\mathsf{p}\in \mathsf{TP}_{\mathsf{a}_\ell}$, a TP-GMM $\boldsymbol{\gamma}_{\mathsf{p}}$ is learned by using $\{\mathsf{TP}_{\mathsf{a}_\ell, \mathsf{p}}\}$ as observations and $\{(\mathbf{s}_j, \mathbf{s}_G)\}$ as the associated frames;
(II) For each object $\mathsf{o} \in \mathsf{O}_{\mathsf{a}_\ell}$, a TP-GMM $\boldsymbol{\gamma}_{\mathsf{o}}$ is learned by using $\{\mathbf{s}_{j, \mathsf{o}}\}$ as observations and $\{(\mathbf{s}_{j, \sim \mathsf{o}}, \mathsf{TP}_{\mathsf{a}_\ell}, \mathbf{s}_G)\}$ as the associated frames, 
where $\mathbf{s}_{j, \sim \mathsf{o}}$ are the states of objects in $\mathsf{O}_{\mathsf{a}_\ell}$ other than~$\mathsf{o}$.
The frames from $\mathbf{s}_j$ and $\mathbf{s}_G$ are mapped to the same set of objects, but instantiated with different system states.
Intuitively speaking, $\{\boldsymbol{\gamma}_{\mathsf{p}}\}$ models how the chosen parameters are constrained w.r.t. the current state and the goal state, while $\{\boldsymbol{\gamma}_{\mathsf{o}}\}$ models how the current state are constrained w.r.t. the chosen parameters and the goal state.
Furthermore, the number of components for \emph{each} TP-GMM within $f_e$ is set to the number of \emph{distinctive} discrete plans from $\boldsymbol{\Xi}$ that contain this edge $e\in E$, i.e., 
\begin{equation}\label{eq:number-comp}
N_e = \Big{|} \big{\{}\boldsymbol{\xi} \in \boldsymbol{\Xi}\;|\; e\in \boldsymbol{\xi}_d\big{\}} \Big{|}, 
\end{equation}
where $\boldsymbol{\xi}_d$ is the associated discrete plan. 
This is essential to distinguish different \emph{modes} of the same edge, i.e., the same transition can be used in different parts of the same plan or different plans.
These TP-GMMs can be learned via the iterative EM algorithm, as described in Sec.~\ref{sec:preliminary}.
Thus, the learned $f(\cdot)$ matches the training data optimally.

The above learning procedure is illustrated in Fig.~\ref{fig:learn-gtn} and summarized in Alg.~\ref{alg:offline-learn}, 
which has polynomial complexity to the state dimension and the number of skills.
As mentioned in Sec.~\ref{subsubsec:structure-GTN}, the representation of $f(\cdot)$ and its learning process described above are novel compared with relevant work. 
First, it is parameterized explicitly w.r.t. both the current state and the goal state. 
Then, it provides a partial abstraction of the state space focused on the geometric relations, i.e., instead of an uniform converge via DNNs as in~\cite{huang2019neural, kim2019learning}.


\begin{figure}[t!]
\centering
\includegraphics[width=0.85\linewidth]{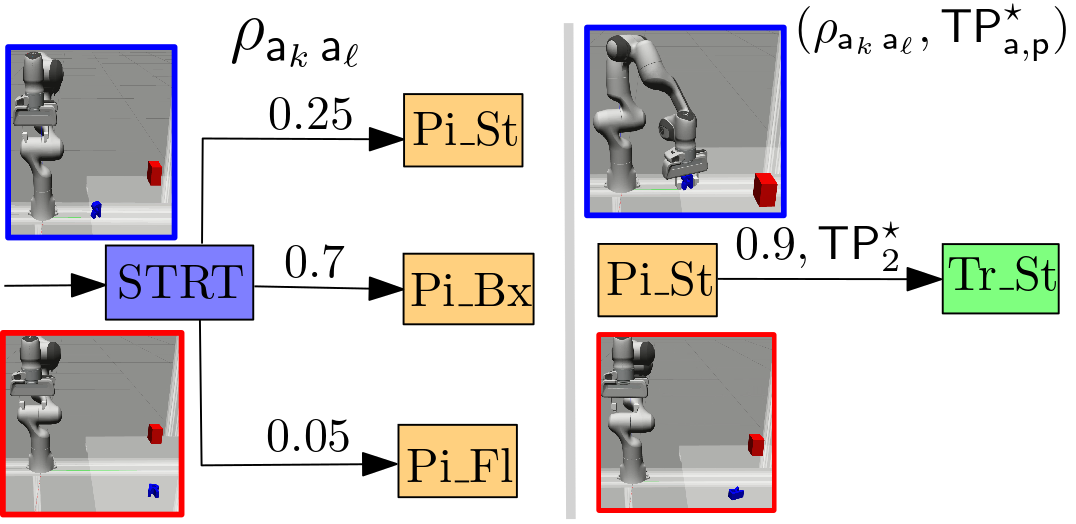}
\caption{
Examples of the transition score from~\eqref{eq:compute-obs-prob}.
{Given different pairs of the initial state $\mathbf{s}_0$ (in blue) and the goal state $\mathbf{s}_G$ (in red), the \textbf{left} figure shows one $\rho_{\mathsf{a}_k \mathsf{a}_\ell}$ (normalized) for various picking skills, 
while the \textbf{right} shows one optimal choice of $\mathsf{TP}^\star$ and  $\rho_{\mathsf{a}_k \mathsf{a}_\ell}$ for different transition.}}
\label{fig:execute-gtn}
\end{figure}

\subsection{Online Execution}\label{subsec:execution-GTN}

This section describes how the learned GTN can be used during online execution to accomplish a new problem of the same task:
particularly, (I) to generate the sequence of skills and the associated parameters;
and (II) to reproduce and execute the skill trajectories. 
At last, we show how failures during execution can be handled. 

\subsubsection{Optimize Skill Sequence and Parameters.}\label{subsubsec:skill-sequence}
Consider that the \emph{new} given problem is to reach~$\mathbf{s}_G$ from~$\mathbf{s}_0$.
Starting from the virtual start node $\underline{\mathsf{a}}$, let $\mathsf{a}_k = \underline{\mathsf{a}}$ and $\mathbf{s}_k = \mathbf{s}_0$. 
For each outgoing transition $(\mathsf{a}_k, \mathsf{a}_\ell) \in E$, 
the following two steps are performed. 
(I) For each \emph{free} task parameter $\mathsf{p}\in \mathsf{TP}_{\mathsf{a}_\ell}$, 
its optimal choice for this transition is given by:
\begin{equation}\label{eq:compute-opt-tp}
\mathsf{TP}^\star_{\mathsf{a}_{\ell}, \mathsf{p}} = \text{argmax}_{\boldsymbol{p}_{\mathsf{a}_{\ell}, \mathsf{p}}}\;\big{\{}\textbf{pdf}\big{(}\boldsymbol{p}_{\mathsf{a}_{\ell}, \mathsf{p}}\,|\,\boldsymbol{\gamma}_{\mathsf{p}}(\mathbf{s}_k,\, \mathbf{s}_G)\big{)}\big{\}},
\end{equation}
i.e., the mean of the corresponding Gaussian computed from the learned  $\boldsymbol{\gamma}_{\mathsf{p}}$ of which the task parameters are derived from~$(\mathbf{s}_k, \mathbf{s}_G)$ as described in~\eqref{eq:f}. 
(II) Then, the ``score'' of this transition is computed by:
\begin{equation}\label{eq:compute-obs-prob}
\begin{split}
\rho_{\mathsf{a}_k \mathsf{a}_\ell}(&\mathbf{s}_k, \mathbf{s}_G)\; \propto \;
  \textbf{Harmonic-Mean}\Big{(}\\ &\big{\{} \textbf{pdf}\big{(}\mathbf{s}_{k, \mathsf{o}}\,|\,\boldsymbol{\gamma}_{\mathsf{o}}(\mathbf{s}_{k,\sim \mathsf{o}},\,  \mathsf{TP}_{\mathsf{a}_{\ell}}^\star,\, \mathbf{s}_G)\big{)},\,\mathsf{o}\in \mathsf{O}_{\mathsf{a}_\ell}\big{\}}\Big{)},
\end{split}
\end{equation}
where \textbf{Harmonic-Mean} is the harmonic mean function;
\textbf{pdf} is the probability density function of a multivariate Gaussian distribution;
$\mathbf{s}_{k, \mathsf{o}}$ is the observation of the current state of object~$\mathsf{o}$;
and the corresponding Gaussian is computed from the learned TP-GMM $\boldsymbol{\gamma}_{\mathsf{o}}$ of which the task parameters are derived from~$(\mathbf{s}_{k,\sim \mathsf{o}},\,  \mathsf{TP}_{\mathsf{a}_{\ell}}^\star, \mathbf{s}_G)$ as described in~\eqref{eq:f};
$\mathsf{TP}_{\mathsf{a}_{\ell}}^\star$ are the optimal choices of parameters for skill $\mathsf{a}_{\ell}$ from~\eqref{eq:compute-opt-tp} above. 
Note that the Harmonic-Mean function takes into account \emph{all} objects within $\mathsf{O}_{\mathsf{a}_\ell}$, instead of choosing only one dominant frame.
Thus, if {any} object is significantly different from the learned constraints, the overall score to that skill is low. 
Fig.~\ref{fig:learn-gtn} illustrates the data flow while computing $\rho_e$.

Consequently, once~$\rho_{\mathsf{a}_k \mathsf{a}_\ell}$ is computed for all possible outgoing transitions $(\mathsf{a}_{k},\mathsf{a}_{\ell})\in E$, 
the next skill to execute is chosen greedily by the computed scores, i.e., 
\begin{equation}\label{eq:opt-a}
\mathsf{a}^\star = \text{argmax}_{\mathsf{a}_\ell}\,\{\rho_{\mathsf{a}_k \mathsf{a}_\ell}\}, 
\end{equation}
i.e., the transition that has the maximum score. 
In this way, the GTN explores during execution \emph{only} the states and skills within the desired plan for the given task.
Note that related work such as~\cite{niekum2015learning, konidaris2012robot, huang2019neural} learns some transition scores similar to~\eqref{eq:compute-obs-prob}, 
but the associated skill parameters are implicitly abstracted from the system state only. 
Instead, the two-step optimization process above allows \emph{explicit} reasoning on the choice of such parameters given the system and goal states, as illustrated in Fig.~\ref{fig:execute-gtn}.

\begin{figure}[t!]
\centering
\includegraphics[width=0.85\linewidth]{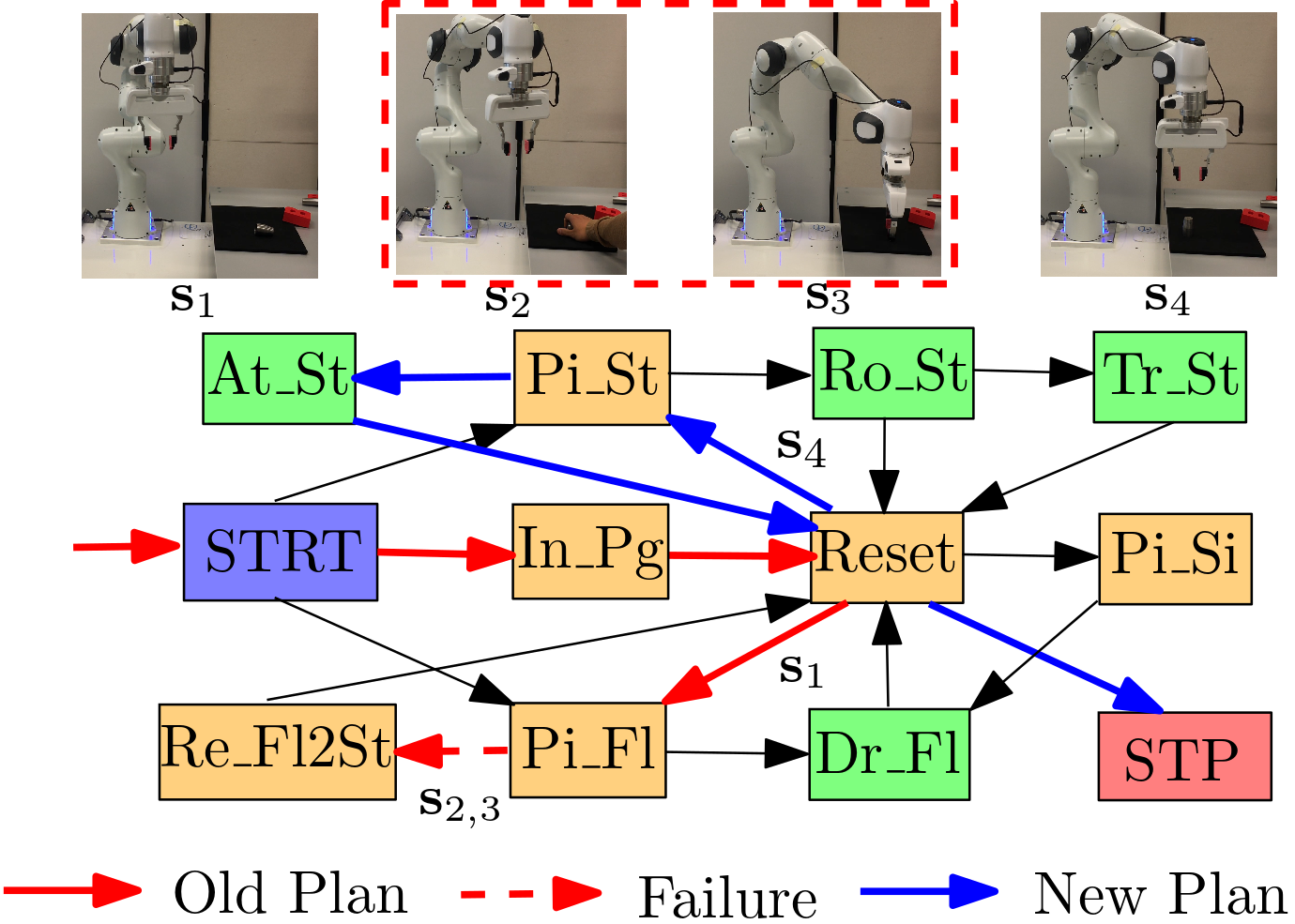}
\caption{
{An example of online failure recovery in Sec.~\ref{subsubsec:failure-recovery}.}
A fault is introduced manually at state $\mathbf{s}_{2,3}$ while picking the ``lying'' cap. 
Afterwards, given the new state $\mathbf{s}_4$, the adapted plan picks the ``standing'' cap again.
}
\label{fig:failure-gtn}
\end{figure}


\begin{algorithm}[t]
   \caption{Online Execution of GTN} \label{alg:online-execution}
   \LinesNumbered
   \DontPrintSemicolon
   \KwIn{$\mathcal{G}=(V, E, f)$, $(\mathbf{s}_0,\mathbf{s}_G)$, $\mathbf{s}_t$.}
   \KwOut{$\{(\mathsf{a}_k, \mathsf{TP}_{\mathsf{a}_k})\}.$}
   Set $\mathsf{a}_k \leftarrow\underline{\mathsf{a}}$ and $\mathbf{s}_k \leftarrow\mathbf{s}_0$.\;
   \While{$\mathbf{s}_k \neq \mathbf{s}_G$}{
       \ForAll{$(\mathsf{a}_k,\,\mathsf{a}_\ell)\in E$}{
           Compute $\mathsf{TP}^\star_{\mathsf{a}_\ell}$ by~\eqref{eq:compute-opt-tp} given $f$ and $(\mathbf{s}_k, \mathbf{s}_G)$.\;
           Compute $\rho_{\mathsf{a}_k \mathsf{a}_\ell}$ by~\eqref{eq:compute-obs-prob} given $f$, $\mathsf{TP}^\star_{\mathsf{a}_\ell}$, $(\mathbf{s}_k, \mathbf{s}_G)$.\;
       }
       Execute $\mathsf{a}^\star$ from~\eqref{eq:opt-a} with parameters $\mathsf{TP}^\star_{\mathsf{a}^\star}$.\;
       Receive new state $\mathbf{s}_t$ from perception.\;
       Set $\mathbf{s}_k\leftarrow \mathbf{s}_t$ and  $\mathsf{a}_k \leftarrow \mathsf{a}^\star$.\;
   }
   \vspace{-0.1cm}
\end{algorithm}

\subsubsection{Skill Execution.}\label{subsubsec:skill-execution}
Once the optimal skill $\mathsf{a}^\star$ is chosen along with its parameter $\mathsf{TP}^\star$,
its trajectory can be retrieved given the previously-learned model~$\boldsymbol{\theta}_{\mathsf{a}}$. 
The retrieval process consists of two steps: First, 
given a partial sequence of observed past trajectory, the most-likely \emph{future} sequence of components $\boldsymbol{k}^\star$ and their durations can be computed via the Viterbi algorithm for classic HSMMs; 
Then, a reference trajectory is generated by a control algorithm (e.g., LQG from~\cite{sciavicco2012modelling}) to track this sequence of Gaussians in the Cartesian space, which is then sent to the motion controller to track, e.g., in the joint space.
More details are given in our earlier work~\cite{rozo2020learning}. 
Afterwards, the system state is changed to $\mathbf{s}_t$, obtained from  the  state estimation and perception modules. 
Given this new state, the same process from Sec.~\ref{subsubsec:skill-sequence} is followed to choose the next skill and its parameters. 
This procedure is repeated until the goal state is reached.

\subsubsection{Failure Recovery.}\label{subsubsec:failure-recovery}
Clearly, if all chosen skills are executed successfully along the plan as described above, the system state should also evolve following the learned GTN structure. 
Namely, after executing skill~$\mathsf{a}_k$, the resulting system state $\mathbf{s}_k$ should satisfy the constraints associated to \emph{at least} one outgoing transition from $\mathsf{a}_k$, i.e., 
\begin{equation}\label{eq:prob-bound}
\rho_{\mathsf{a}_k \mathsf{a}_\ell}(\mathbf{s}_k, \mathbf{s}_G) >\underline{\rho}, \quad \exists (\mathsf{a}_k, \mathsf{a}_\ell) \in E,
\end{equation}
where $\underline{\rho}>0$ is a manually-chosen lower bound on the computed transition score from~\eqref{eq:compute-obs-prob}.
If the condition in~\eqref{eq:prob-bound} is not satisfied, it indicates that the system does \emph{not} evolves as learned, and none of the following skills are plausible. 
After manually checking the execution, two measures can be taken: 
(I) if skill~$\mathsf{a}_k$ is executed successfully, it indicates that the learned skill model~$\mathcal{M}_{\mathsf{a}_k}$ can not cover this scenario.
Then additional demonstrations can be provided that are close to this scenario. 
(II) if skill~$\mathsf{a}_k$ is not executed successfully due to uncertainties, e.g., in perception and motion.
Then, the current progress should be re-identified for the given task, as a failure could potentially set the progress forward or backward.
This is done by evaluating the transition scores of \emph{all} edges and choose the one with the highest score, namely: 
\begin{equation}\label{eq:adapt}
(\mathsf{a}_{m},\, \mathsf{a}_n)^\star = \text{argmax}_{\mathsf{a}_{m}\mathsf{a}_n \in E} \; \big{\{}\rho_{\mathsf{a}_{m}\mathsf{a}_n}(\mathbf{s}_k, \mathbf{s}_G)\big{\}},
\end{equation}
which is an expensive process compared with~\eqref{eq:opt-a}.
Note that if the condition in~\eqref{eq:prob-bound} is still not satisfied by $(\mathsf{a}_{m},\, \mathsf{a}_n)^\star$, then the system has entered a \emph{unrecoverable} state and has to be reset.
An example is shown in Fig.~\ref{fig:failure-gtn}.

\subsubsection{Algorithmic Summary.}\label{subsubsec:algorithmic-summary}
The components described above are summarized in Alg.~\ref{alg:offline-learn} for offline learning and in Alg.~\ref{alg:online-execution} for online execution. 
Note that the \emph{only} inputs required are the demonstrations $\{\mathsf{D}_{\mathsf{a}}\}$ and the task specifications as $\mathbf{s}_0$ and $\{\mathbf{s}_G\}$.
The TAMP solver needed for the data generation can be constructed as described in Sec.~\ref{subsubsec:train-data}.
During online execution, access to the perception and control modulars are required. 
More implementation details are given in Sec.~\ref{sec:experiments}.

Last but not least, it is worth stressing that the policy during online execution is fully \emph{explainable}, regarding: 
(I) why the policy chooses certain skill and these parameters,
(II) what are the alternatives, 
(III) how confident such choices are, 
and (IV) how to detect failures and how the system could recover.
In particular, regarding (I)-(III), \eqref{eq:compute-opt-tp} provides insights on how each frame associated with current state or goal state contributes to the choice of optimal $\mathsf{TP}_{\mathsf{a}_\ell, \mathsf{p}}^\star$, 
while~\eqref{eq:compute-obs-prob} and~\eqref{eq:opt-a} indicate how each object state contributes to the confidence of next skills.
The recovery mechanism by~\eqref{eq:prob-bound} and~\eqref{eq:adapt} reveals the underlying reasoning regarding (IV).
This leads to better transparency with human operators during the process.

{One variation of the proposed GTN is by grouping similar primitive skills into one general skill, 
e.g., by combining ``pick stand'' and ``pick flat''  into one ``pick'' skill.
Even though this would result in a simpler GTN, 
this does not reduce the overall complexity as the choice of sub-skills is transferred \emph{inside} each general skill. 
On the contrary, this would make the GTN less transparent regarding how such choice depends on the goal.}

\begin{figure}[t!]
\centering
\includegraphics[width=0.9\linewidth]{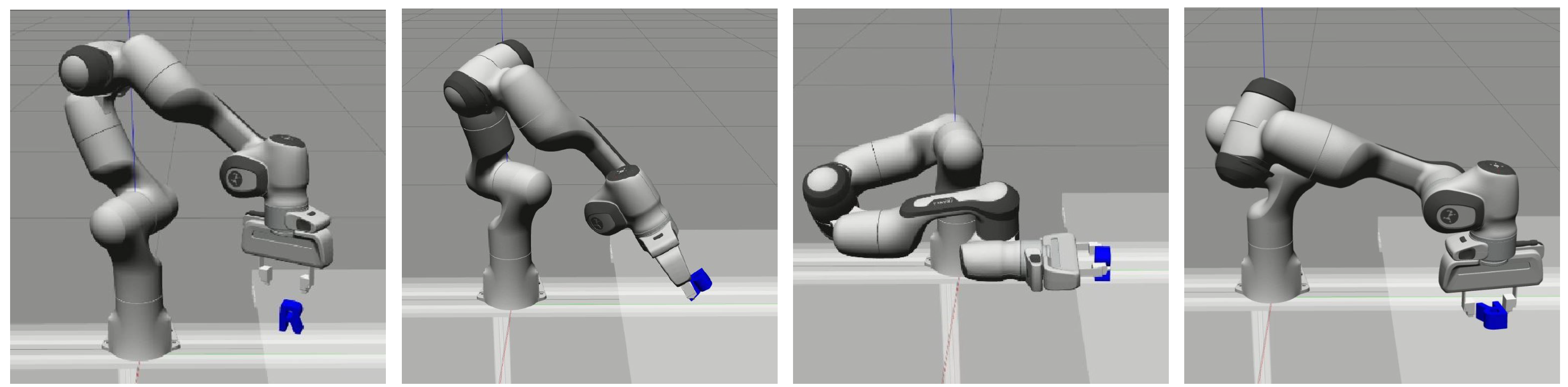}

\includegraphics[width=0.85\textwidth]{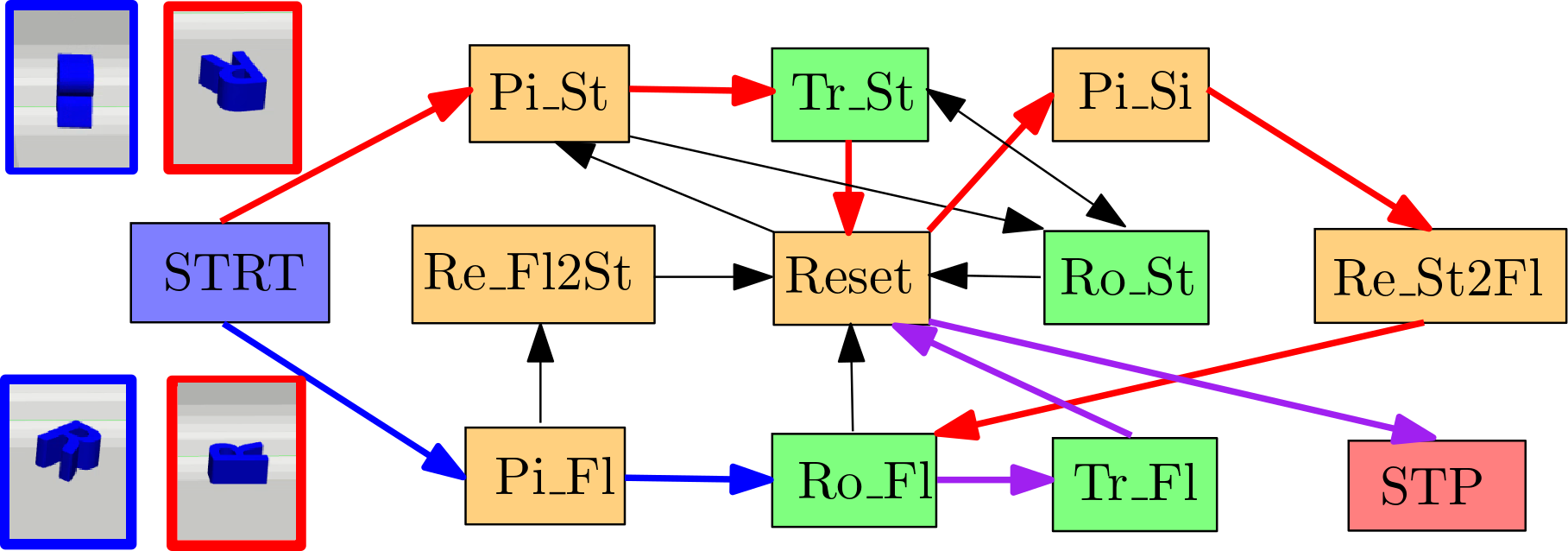}
\caption{
Top (from left to right): snapshots of skills \texttt{Pi\_St}, \texttt{Re\_St2Fl}, \texttt{Pi\_Si} and \texttt{Ro\_Fl}. 
Bottom: the learned GTN associated \textbf{Task-1}, where two different plans are highlighted (in blue and in red, shared edges in purple).
}
\label{fig:task-1}
\end{figure}

\section{Experiments}\label{sec:experiments}

This section contains the numerical validation over a 7-DoF robotic manipulator, both in simulation and on actual hardware.
The proposed approach is implemented in Python3 on top of Robot Operating System (ROS) to enable communication across planning, control and perception modules.
All benchmarks are run on a laptop with 8-core Intel Xeon CPU.
Detailed descriptions of the simulation and hardware setup, and experiment videos can be found in the supplementary file.

\subsection{6D Scrabble in Simulation}\label{subsec:sim}
\begin{figure}[t!]
\centering
\includegraphics[width=0.9\linewidth]{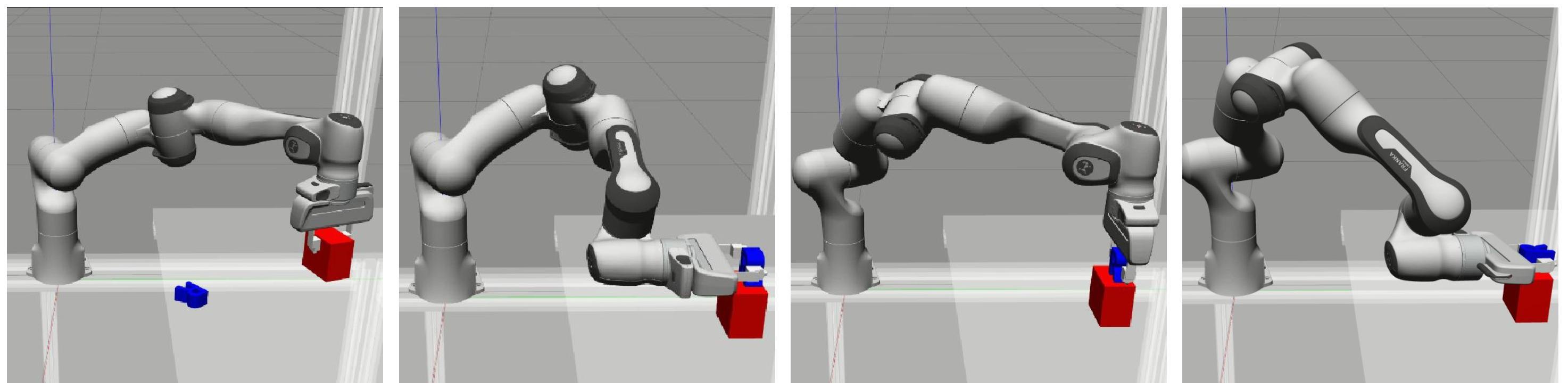}

\includegraphics[width=0.85\textwidth]{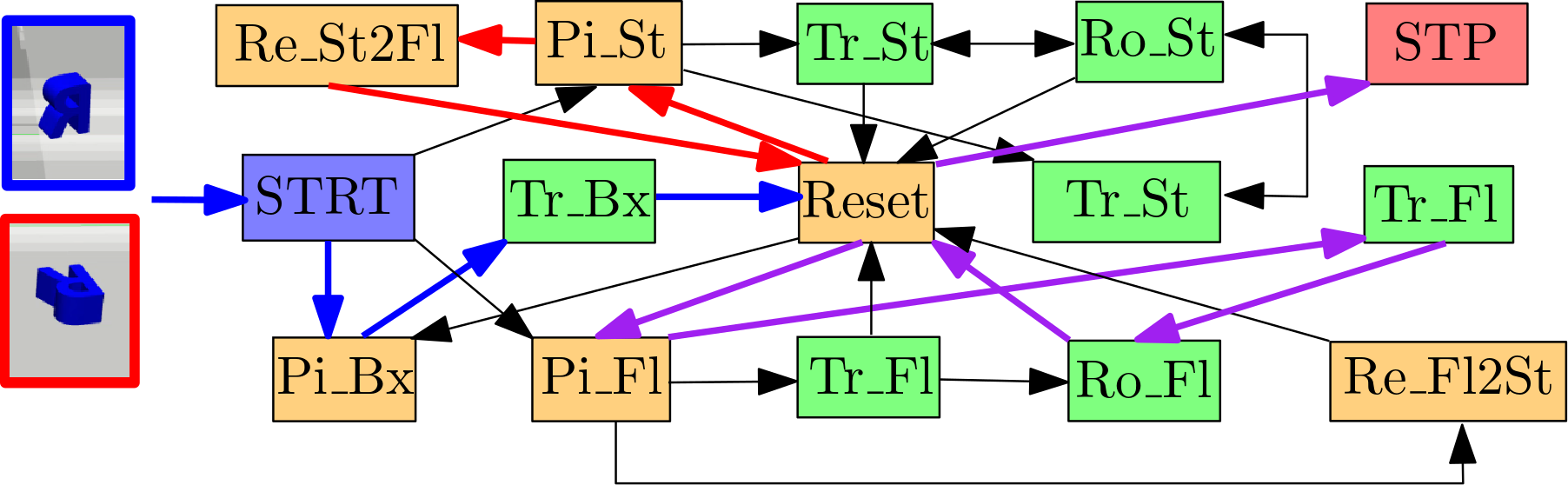}
\caption{
Top (from left to right): snapshots of skills \texttt{Pi\_Bx}, \texttt{Re\_Fl2St} (onto the box), \texttt{Tr\_St}  and \texttt{Re\_St2Fl} (onto the box). 
Bottom: the learned GTN for \textbf{Task-2}, where one plan is highlighted in blue, red, and purple. 
}
\label{fig:task-2}
\end{figure}

To avoid limitations from perception, we consider first in simulation three manipulation tasks with increasing complexity. 
Particularly, several 3D letters are scattered on a platform along with a 7-DoF Franka arm, as shown in Fig.~\ref{fig:task-1},~\ref{fig:task-2},~\ref{fig:task-3}.

\textbf{Task-1}: manipulate one {letter} to reach the desired pose in 6D. 
Note that re-orientation and rotation skills are used to change its orientation (in pitch and in yaw).
Moreover, different picking skills are required for grasping the object at different orientations. 
This is different from symmetric objects such as blocks in~\cite{huang2019neural} where only ``pick-and-place'' skills are sufficient.
The purpose of this task is to illustrate the computational complexity even for a simple task.

\textbf{Task-2}: similar to \textbf{Task-1} but with a box to levitate the {letter} for re-orientation, instead of at the platform edge. 
Note that this subtle difference makes this task more difficult as the box is not directly related to the goal but used as a tool to facilitate other subsequent skills. 

\textbf{Task-3}: manipulate \emph{three} {letters} with a levitation box to a desired formation.
This task shows the increased complexity when there are more relevant objects, to emphasize the \emph{order} of manipulating different objects and using a tool in between.

Note that the state-space of tasks above is high-dimensional, 
e.g., 6D pose of the end effector and all objects, 
which are well-known to be difficult for classic TAMP methods.

\begin{table}[!t]
\caption{
Models of skills used in \textbf{Task-1,2,3}.}
\label{table:skill}
    \begin{center}
    \begin{threeparttable}
    \begin{adjustbox}{width=0.85\linewidth}
        \begin{tabular}{c||ccccc}
        \toprule
        Skill Name & $M_{\mathsf{a}}$ & $K_{\mathsf{a}}$  &  $\mathsf{TP}_{\mathsf{a}}$ & $t\,(\boldsymbol{\theta}_{\mathsf{a}}\,\vert \, \boldsymbol{\gamma}_{\mathsf{a}})$ [\si{\second}]\\ \midrule
        \texttt{Pi\_St} & $12$ & $7$ &  $\{{A}, {R}\}$ & $3.6\, \vert\,  2.7 $\\
        \texttt{Pi\_Si} & $10$ & $8$ &  $\{{A}, {R}\}$ & $4.2\, \vert\,  3.1 $\\
        \texttt{Pi\_Fl} & $8$ & $7$ &  $\{{A}, {R}\}$ & $3.8\, \vert\,  3.2 $\\
        \texttt{Re\_St2Fl} & $9$ & $8$ &  $\{{A}, {R}, {A}_G\}$ & $4.3\, \vert\,  2.1 $\\
        \texttt{Re\_Fl2St} & $10$ & $8$ &  $\{{A}, {R}, {A}_G\}$ & $3.8\, \vert\,  2.2 $\\
        \texttt{Ro\_St} & $15$ & $6$ &  $\{{A}, {R}, {A}_G\}$ & $4.2\, \vert\,  2.7 $\\
        \texttt{Ro\_Fl} & $10$ & $6$ &  $\{{A}, {R}, {A}_G\}$ & $3.3\, \vert\,  2.2 $\\
        \texttt{Tr\_St} & $10$ & $7$ &  $\{{A}, {R}, {A}_G\}$ & $4.1\, \vert\,  1.9 $\\
        \texttt{Tr\_Fl} & $10$ & $7$ &  $\{{A}, {R}, {A}_G\}$ & $4.3\, \vert\,  2.1 $\\
        \texttt{Reset} & $7$ & $6$ &  $\{{R}\}$ & $2.8\, \vert\,  1.5 $\\
        \bottomrule
        \end{tabular}
    \end{adjustbox}
    \end{threeparttable}
    \end{center}
\end{table}

\subsubsection{Primitive Skills}\label{subsec:sim-skills}
For each {letter}, there are~$10$ primitive skills relevant to the above tasks:
\texttt{Pi\_St} and \texttt{Pi\_Si} to pick a standing letter from the top or the side 
while \texttt{Pi\_Fl} to pick a flat-lying letter;
\texttt{Re\_St2Fl} re-orients a letter from standing to lying (i.e., its pitch angle),  while \texttt{Re\_Fl2St} does the opposite;
\texttt{Ro\_St} and \texttt{Ro\_Fl} rotate a letter by \emph{arbitrary} yaw angle while standing and flat, respectively; 
\texttt{Tr\_St} and \texttt{Tr\_Fl} translate a letter to \emph{arbitrary} position without changing its orientation;
\texttt{Reset} resets the robot.
Instead of kinesthetic teaching, a 6D mouse is used to generate demonstrations for each skill under various configuration of the robot and the letters, recorded at \SI{10}{\hertz} within the simulator. 
As summarized in Table~\ref{table:skill}, 
{in average~$9$ demonstrations are performed for each skill, due to a large workspace and various objects}.
The associated trajectory model $\boldsymbol{\theta}_{\mathsf{a}}$ is learned in \SI{4}{\second} with around $8$ components and two or three frames, 
while the condition-effect model $\boldsymbol{\gamma}_{\mathsf{a}}$ is learned in \SI{3}{\second}.

\subsubsection{Results for Task-1}\label{subsec:sim-results-task-1}

Starting from \textbf{Task-1}, Alg.~\ref{alg:offline-learn} is followed to learn the associated GTN.
More specifically, the training set consists of $100$ problems that are generated by randomly choosing pairs of initial and final poses for the letter.
It took in total~\SI{7}{\hour} the proposed hybrid TAMP to solve all problems successfully in the training set.
Afterwards, the associated GTN is learned in \SI{2.8}{\second} of which the constraints $f_{\texttt{e}}$ takes around~\SI{0.1}{\second} to compute for each edge. 
As shown in Fig.~\ref{fig:task-1}, the learned GTN has $12$ nodes, $20$ edges and $28$ TP-GMMs for the constraints, which encapsulates all $4$ possible plans for \textbf{Task-1}, which have length $4$, $4$, $7$ or $9$ and contain $2$, $2$, $2$ and $4$ skill parameters (\emph{each} as 6D pose).
Namely, the desired behavior is already quite complex: 
if the initial and goal pitch angle of the letter are the same (i.e., both standing or lying), then the robot needs to pick the {letter} with the correct pick skill, rotate it to the desired yaw angle and translate it to the desired position.
However, if they are different (i.e., from standing to lying or vice versa), then re-orientation is needed.
Note that skill \texttt{Re\_St2Fl} requires the {letter} to be grasped from the side via skill \texttt{Pi\_Si}, which is \emph{only} allowed at the platform edge to avoid close contact between the robot and the platform. 
Afterwards, the learned GTN is used to solve $100$ new problems for validation by Alg.~\ref{alg:online-execution}. 
It takes around \SI{0.7}{\second} per query to GTN and \SI{5}{\second} per problem with a success rate of $100\%$, resulting in a $100$-fold decrease in planning time compared with the TAMP solver.
{The derived plan and the associated skill trajectories are tracked using
the onboard impedance controller to compute appropriate joint torques.}

\begin{figure}[t!]
\centering
\includegraphics[width=0.9\linewidth]{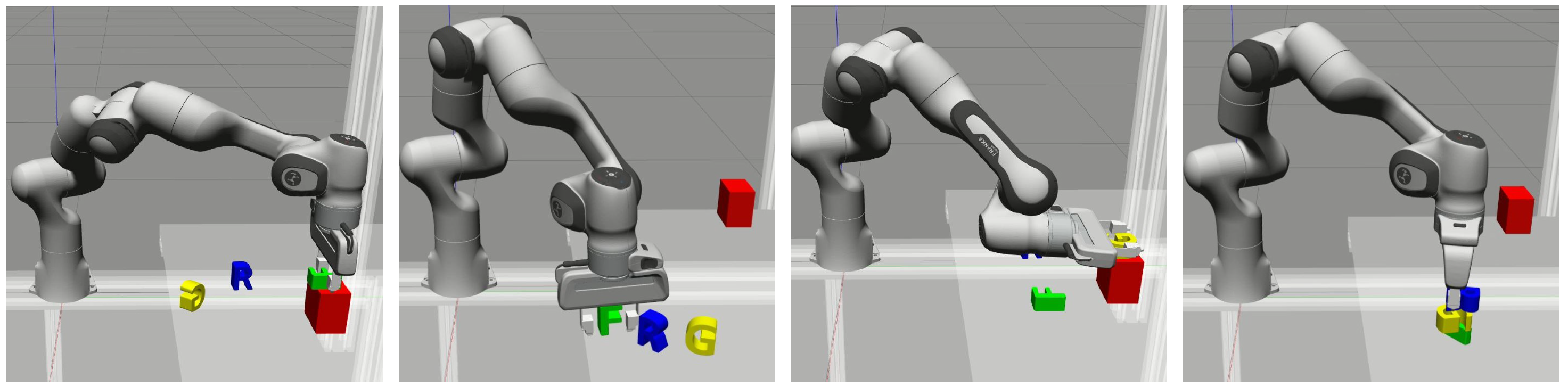}

\includegraphics[width=0.85\textwidth]{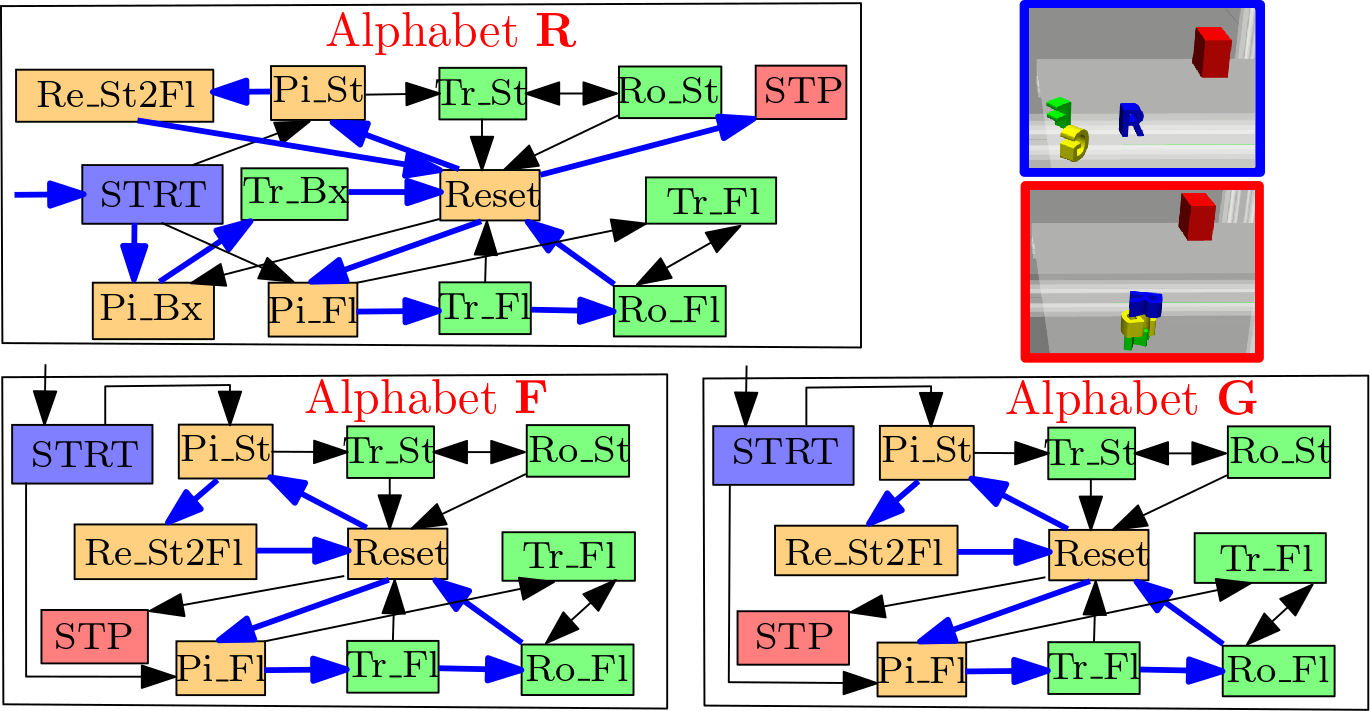}
\caption{
Top (from left to right): snapshots of skills~\texttt{Pi\_Fl} (from the box), \texttt{Ro\_St}, \texttt{Re\_St2Fl} (onto the box), 
and \texttt{Tr\_Fl}. 
Bottom: the learned GTN associated with \textbf{Task-3}, where one plan is highlighted in blue. 
Note the ``start'' and ``stop'' nodes for each letter is duplicated for the ease of visualization.
}
\label{fig:task-3}
\end{figure}
%

\subsubsection{Results for Task-2}\label{subsec:sim-results-task-2}

Furthermore, \textbf{Task-2} has the same task description as before but a levitation box is introduced as a new object. 
This box can be used to facilitate the re-orientation of the {letter} from the platform to the box top after being grasped.
Note that to increase the task difficulty, re-orientation of the letter on the platform edge as in {\textbf{Task-1}} is disabled in this case, but instead on the box top.
Moreover, the box should be moved to an optimal pose beforehand to facilitate the re-orientation, which can \emph{not} be inferred directly from the goal state (which actually indicates the box should not be moved). 
The same procedure as before is followed to learn the associated GTN.
In total, $150$ problems are generated for training and solved successfully by the same TAMP solver within~\SI{18}{\hour}.
Afterwards, the associated GTN is learned in \SI{7.8}{\second}, which has $15$ nodes and $27$ edges, as shown in Fig.~\ref{fig:task-2}. 
It  encapsulates all $4$ possible plans with length $4$, $4$, $12$ or $14$ and $2$, $2$, $4$ and~$6$ 6D parameters. 
The time for generating training data is much longer due to the larger number of skills and higher dimension of parameter space. 
Note that the box is only picked and translated when the initial and goal states are not aligned in pitch angle, and then moved back afterwards. 
Then the learned GTN is validated against a set of $150$ problems, which take in average \SI{0.8}{\second} per query to GTN and \SI{8}{\second} per problem with a success rate of $96\%$ (compared with  \SI{7}{\minute} per problem during the generation of training data).

\begin{table}[!t]
\caption{
Model size and computation time for all tasks.
}
\label{table:results-all}
    \begin{center}
    \begin{adjustbox}{width=0.9\linewidth}
      \begin{tabular}{c||cccccc}
	\toprule
        Task           & $\boldsymbol{\Xi}$ & $t_D\, [\SI{}{\hour}]$ & $|\mathcal{G}|$ & $t_L\,[\SI{}{\second}]$ & $t_V\,[\SI{}{\second}]$ & $t_E\,[\SI{}{\second}]$\\ \midrule
\\[-10pt]
       \textbf{Task-1} & 100      & 7.3         & (12, 20, 28)      & 2.8       & 5.8     & 0.7            \\
\\[-8pt]
       \textbf{Task-2} & 150      & 18.1         & (15, 27, 32)      & 7.8       & 8.9    & 0.8             \\
\\[-8pt]
       \textbf{Task-3} & 500      & 101         & (29, 53, 256)      & 245       & 19.5   & 1.5              \\
\\[-8pt]
       \textbf{Assembly} & 80      & 2.5         & (12, 19, 27)      & 1.3       & 3.8    &  0.3            \\
	\bottomrule
      \end{tabular}
    \end{adjustbox}
\parbox{\textwidth}{\footnotesize %
\vspace{1eX} 
Training data size $\boldsymbol{\Xi}$, the time to generate such data~$t_D$, the size of GTN $|\mathcal{G}|$, the learning time $t_L$, the total solution time of the validation set (the same size as $\boldsymbol{\Xi}$), and the average evaluation time $t_E$ to output $(\mathsf{a}^\star, \,\mathsf{TP}^\star)$.}
    \end{center}
\end{table}

%
\subsubsection{Results for Task-3}\label{subsec:sim-results-task-3}
As the most complex task, \textbf{Task-3} involves $4$ objects where $3$ letters should be re-arranged according to the goal state with the box being a tool (as in \textbf{Task-2}). 
Two aspects of the task are challenging: 
(I) the sequence of manipulating these letters is critical when the letters are stacked as the goal; 
and (II) the box can be shared as a tool for multiple letters. 
In total $500$ problems are generated for training and equally distributed for different cases such as distributed and stacking.
All problems are solved successfully by the proposed hybrid TAMP solver, but with a drastically longer time of around~\SI{12}{\minute} per problem. 
This is because in this case there are~$29$ potential skills and~$14$ 6D parameters in total to sample at \emph{each} state during the search. 
In addition, an incorrect sequence of manipulating the letters would yield large parts of the search graph invalid. 
Given these solutions, the associated GTN is learned in \SI{4}{\minute}, which has $29$ nodes and $53$ edges as shown in Fig.~\ref{fig:task-3}. 
The learning time is also increased significantly due to fact that there are in total $18$ possible paths with maximum length $31$ and $14$ 6D parameters. 
The constraint function $f$ consists of $256$ TP-GMM components in total, which is $8$ times more than \textbf{Task-2}. 
Clearly, a manual specification of all such constrains and paths would be a tedious process, if not impossible. 
It is interesting to notice that 
(I) when the letters are stacked together in the goal state, the transition score to the picking skills of different {letters} changes according to the order of {letters} in the stack; 
(II) the box is used only when \emph{at least} one letter needs re-orientation, i.e., the score of picking the box first is maximized when the yaw angle is not aligned for at least one {letter}.
During validation against $500$ new problems, the solution time is in average~\SI{20}{\second} per problem and~\SI{1.5}{\second} per edge. 
However the overall success rate is $95\%$, where failures often happens when the final stacking does not satisfy the desired accuracy due to the large force applied  by the robot.

\begin{figure}[t!]
\centering
\includegraphics[width=0.9\linewidth]{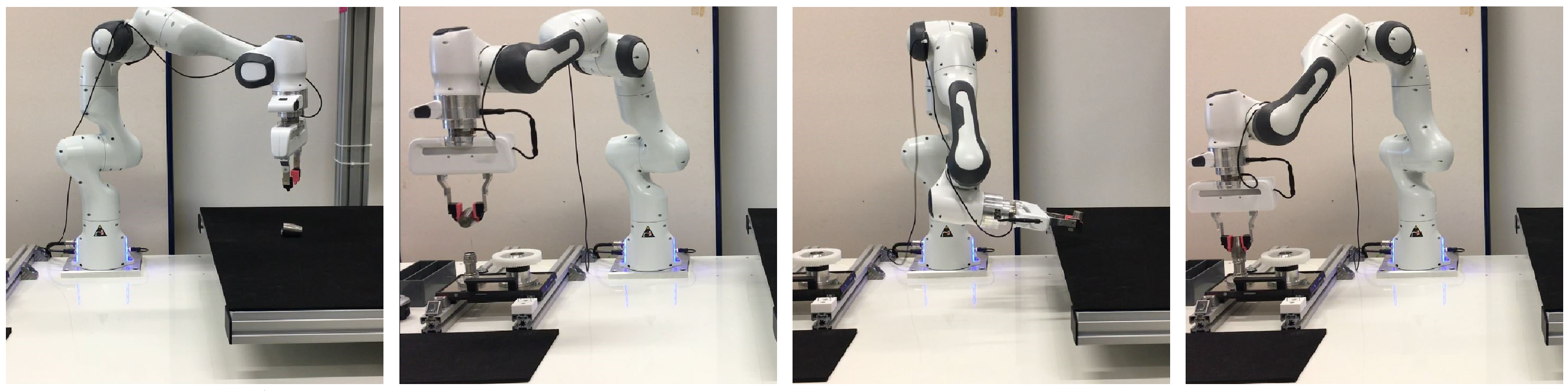}

\includegraphics[width=0.85\textwidth]{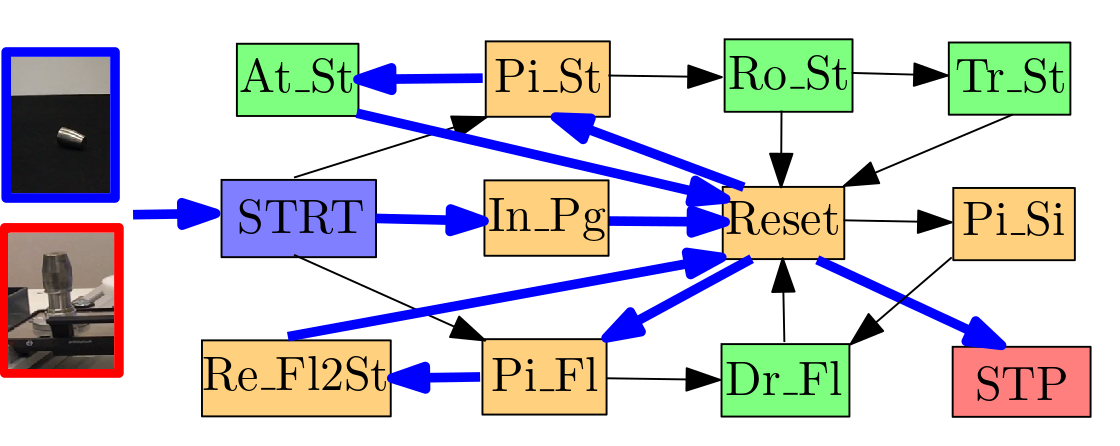}
\caption{
Top (from left to right): snapshots of skills~\texttt{Pi\_Fl}, \texttt{Dr\_Fl}, \texttt{Re\_Fl2St} (on the platform), 
and \texttt{At\_St}. 
Bottom: the learned GTN for \textbf{Assembly-Task}, where one plan is highlighted in blue for the shown problem. 
}
\label{fig:hw}
\end{figure}


\subsection{Industrial Assembly on Hardware}\label{subsec:hw}
To further validate the proposed framework, we consider some steps of an industrial assembly task.  
The actual Franka robotic arm is used in a workspace that consists of a feeding and inspection platform, and an assembly station where various pieces are assembled into a product, as shown in Fig.~\ref{fig:hw}. 
The platform is monitored by a Zivid camera, which provides a 6D pose estimation with less than \SI{1}{\centi\metre} accuracy.
During kinesthetic teaching, the end-effector state is fetched directly from the state estimator. 
Demonstrations are recorded at \SI{10}{\hertz}, while the impedance controller runs at \SI{1}{\kilo\hertz}.

\subsubsection{Task Description.}\label{subsec:hw-task}
The workstation consists of an assembly station, a pallet and a feeding platform, 
while the objects of interest are a metallic cap and a peg (as components of an e-bike motor). 
Since caps are loaded onto the platform, subsequently
(I) If the cap is defective, it should be picked and dropped into the pallet (while lying); 
(II) If the cap is non-defective, the cap should be picked and attached to the top of the peg (while standing), 
which should be inserted into the assembly station first. 
Similar to \textbf{Task-1}, the cap orientation effects greatly the desired skill sequence.

There are in total $10$ skills taught to the robot \emph{directly} via kinesthetic teaching, which are quite similar to those in Sec.~\ref{subsec:sim-skills}. 
Additional skills include \texttt{In\_Pg} to insert the peg into the workstation, \texttt{At\_St} to attach the cap onto the peg while standing, and \texttt{Dr\_Fl} to drop the cap while lying.
Details can be found in the supplementary file.

\subsubsection{Results.}\label{subsec:hw-result}
The goal states are generated based on the whether it should be attached to the peg, or dropped into the pallet. 
In total, $80$ problems are generated for training and solved successfully by the proposed TAMP solver within~\SI{2.5}{\hour}.
Afterwards, the associated GTN is learned in only \SI{1.3}{\second}, which has $12$ nodes,~$19$ edges and~$27$ embedded TP-GMMs, as shown in Fig.~\ref{fig:hw}. 
It encapsulates all $4$ possible plans with length $5$,~$8$,~$3$ and~$7$, with maximum $2$ 6D parameters. 
Namely, if the cap is initially standing but needs to be in the pallet as the goal, it is translated first to the platform edge then grasped from the side;
while if the cap is initially lying but needs to be on the peg as the goal, it is firstly  re-oriented to standing at the platform edge then attached to the peg. 
As summarized in Table.~\ref{table:results-all}, the learned GTN is validated on hardware for new problems, which take in average \SI{0.3}{\second} per query to GTN and \SI{4}{\second} per problem with a success rate of $100\%$ (compared with  \SI{2}{\minute} per problem during data generation).

\begin{table}[t!]
\caption{
Learning time for each method and each task.
}
\label{table:learning-time}
    \begin{center}
    \begin{adjustbox}{width=0.9\linewidth}
      \begin{tabular}{c||ccccc}
	\toprule
        Time [\SI{}{\second}]           & $|\boldsymbol{\Xi}|$ & \textbf{Hb-lfd/mp} & \textbf{GTN}& \textbf{NPI} & \textbf{MLP}\\ \midrule
\\[-10pt]
       \textbf{Task-1} & 100 & N/A      &  28      & 182       & 53                \\
\\[-8pt]
       \textbf{Task-2} & 150 & N/A      & 32      & 250       & 70                \\
\\[-8pt]
       \textbf{Task-3} & 500 & N/A      &  256      & 582      & 120               \\
\\[-8pt]
       \textbf{Assembly} & 80 & N/A    & 27      & 84      & 57                \\
	\bottomrule
      \end{tabular}
      \end{adjustbox}
    \end{center}
\end{table}

\begin{figure}[t!]
\centering
\includegraphics[width=0.98\textwidth]{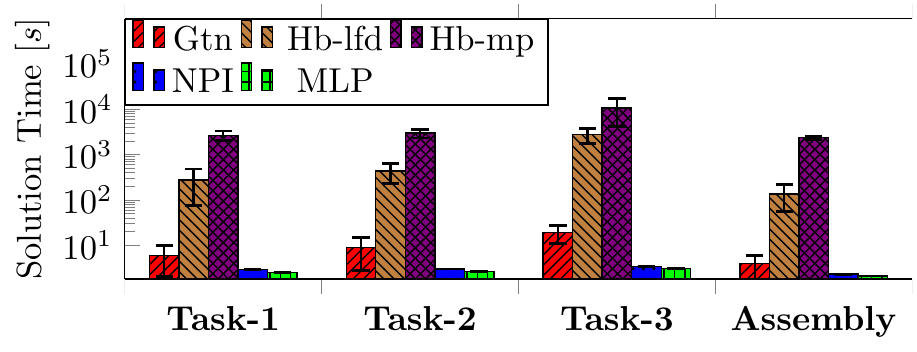}
\caption{
Comparison of the solution time during validation for all tasks, between GTN and four baselines.
}
\label{fig:results-time}
\end{figure}

In addition, the failure recovery mechanism as described in Sec.~\ref{subsubsec:skill-execution} is demonstrated here with two types of failures: 
(I) The cap is removed and re-oriented from ``lying'' to ``standing'' by the operator, after skill \texttt{Pi\_Fl} is started. 
Consequently,  the cap is not grasped and the score of the only outgoing transition to skill \texttt{Re\_Fl2St} is computed as \SI{2e-4}, 
which is lower than the pre-defined threshold $0.1$. 
The failure condition in~\eqref{eq:prob-bound} is satisfied and the adaptation rule in~\eqref{eq:adapt} is triggered.
The edge from \texttt{Tr\_St} to \texttt{Reset} is chosen as the starting edge. 
Afterwards, the system continues to pick and attach the cap. 
(II) The skill \texttt{Re\_Fl2St} is not executed successfully as the cap falls back to lying on the platform. 
The resulting relative transformation between the end-effector and the cap is different from prediction. 
Thus the transition score to skill \texttt{Reset} is computed as \SI{1e-3}, which is identified as a failure by~\eqref{eq:prob-bound}. 
Similar as before, the edge from skill \texttt{Tr\_St} to skill \texttt{Reset} is chosen as the  starting edge. 
Afterwards, the system \emph{repeats} th procedure to pick and reorient the cap.

%
\subsection{Comparison and Discussion}\label{subsec:sim-compare}
{In this section}, we compare the proposed approach against other baselines and discuss further its imitations.

\begin{figure}[t!]
\centering
\includegraphics[width=0.8\textwidth]{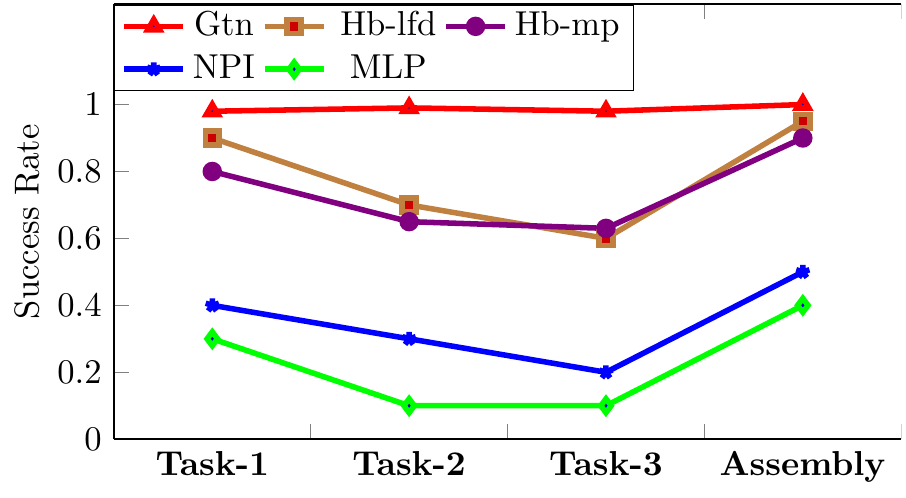}
\caption{
{Success rate of each method for each task during validation.}
}
\label{fig:results-rate}
\end{figure}

\subsubsection{Baselines}\label{subsubsec:baseline}
We consider four baselines below:

\textbf{Hb-lfd}: the hybrid search planner described in Sec.~\ref{subsubsec:train-data} over the learned LfD skills, 
and using the confidence measure from~\cite{Schwenkel2019Optimizing} to compute the cost heuristic.

\textbf{Hb-mp}: the hybrid search planner over the skills generated from sampling-based motion planner RRTC {(see~\cite{lavalle2006planning})}
with the ``distance to goal'' as the cost heuristic. 
This is commonly used when a precise simulator is available with no domain knowledge. 
The simple heuristic prefers the skill that drives the system closer to the goal state during the search.

\textbf{MLP}: a multilayer feedforward neural network to predicate both the next skill and its parameters. 
A 3-layer structure is trained in a supervised way, where the inputs include \emph{both} the system state and the goal state, whether the optimal choice of next skill and the parameters are the outputs.
The loss function is the MSE between the target an the prediction.

\textbf{NPI}: a LSTM-based policy network similar to the Neural Programmer Interpreter from~\cite{huang2019neural}. 
It is proposed to train a network which coordinates programs under different contexts. 
We use one hidden layer with a memory of $5$ steps. 
The output and the loss function remain the same as those in \textbf{MLP}, but the input is now a \emph{sequence} of past system states, goal states and the next chosen skills.

The metrics to compare are (I) the learning time;
(II) the solution time and the success rate during validation. 
Note that the same training and validation sets are used for the above baselines. 
Moreover, both \textbf{NPI} and \textbf{MLP} need to learn directly from the solution set $\boldsymbol{\Xi}$, 
thus the data generation from Sec.~\ref{subsubsec:train-data} remains necessary for both methods.

\subsubsection{Comparison}\label{subsubsec:comparison}
\begin{figure}[t!]
\centering
\includegraphics[width=0.49\textwidth]{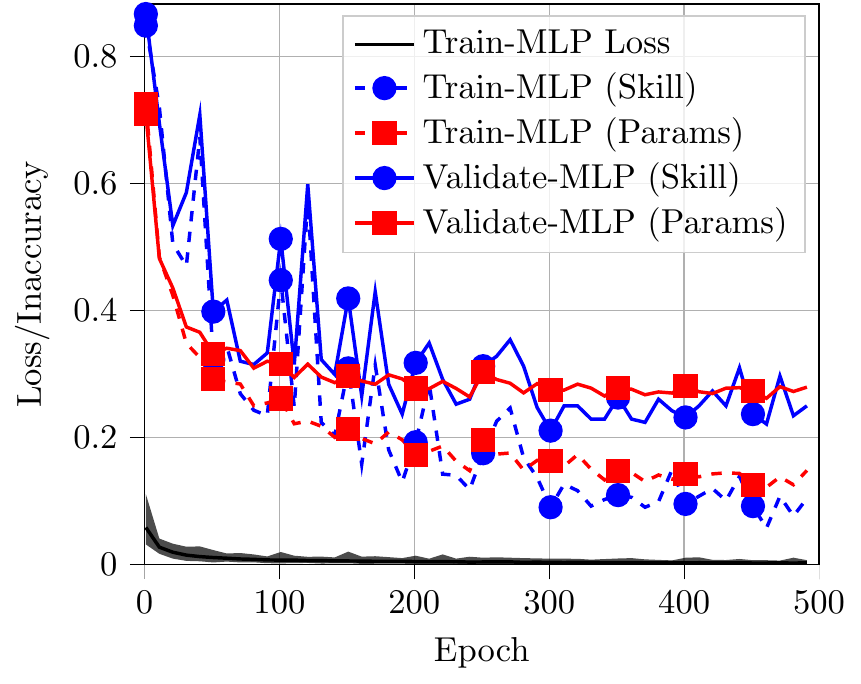}
\includegraphics[width=0.49\textwidth]{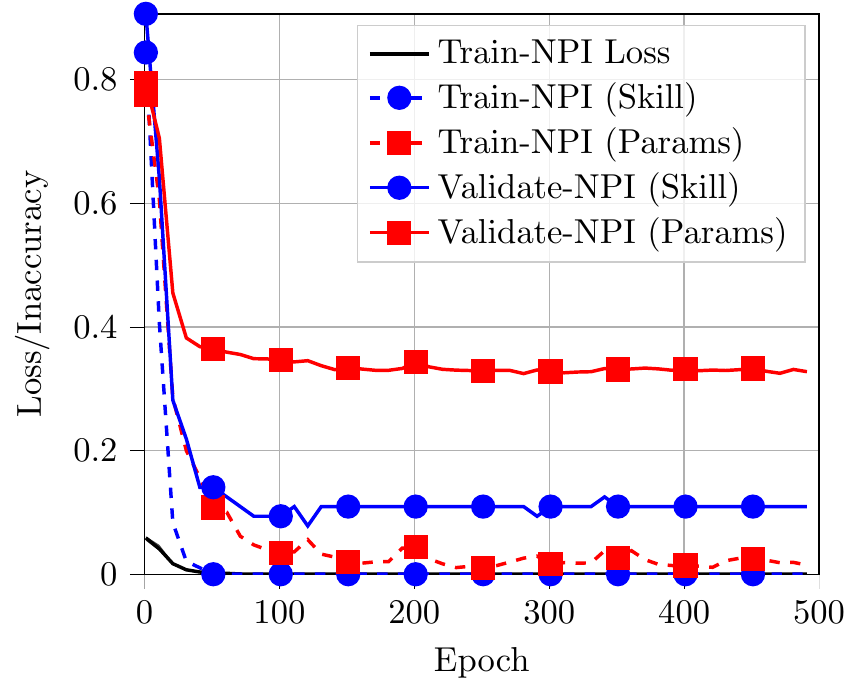}
\caption{
Evolution of training loss for \textbf{MLP} and \textbf{NPI} during learning for \textbf{Task-2}.
Note the difference between the training (dashed) and validation (solid) inaccuracy, for predicting the next skill (in blue) measured by the rate of incorrect choices,  and the associated parameter (in red) measured by the MSE.
}
\label{fig:results-train}
\end{figure}

We notice that these four baseline planners perform quite consistently across all four tasks. 
First, regarding the first metric,  Table.~\ref{table:learning-time} summarizes the learning time for all methods.
In particular, \textbf{Hb-lfd} and \textbf{Hb-mp} do not learn from past solutions. 
The learning time of \textbf{GTN} is  much lower than those needed for \textbf{NPI} and \textbf{MLP} for all four tasks. 
The stopping criteria for both \textbf{NPI} and \textbf{MLP} is that the training loss is below \SI{1e-4} or until $500$ epochs. 
As shown in Fig.~\ref{fig:results-train}, the training loss decreases steadily to zero and prediction accuracy improves till a steady state.

Second, these four baselines are evaluated against the same validation set. 
As summarized in Fig.~\ref{fig:results-time}, \textbf{GTN} shows a $100$-fold \emph{decrease} in solution time to around \SI{10}{\second} for all tasks compared with the search planners. 
{Notably, \textbf{Hb-mp} take the longest as it needs to query the simulator constantly to assess system evolutions, execution outcome and cost} (around \SI{7}{\second} per query), while \textbf{Hb-lfd} is $10$-times faster via the learned condition and effect model. 
Furthermore, the simple heuristic in  \textbf{Hb-mp} is not useful, e.g., the box usage in \textbf{Task-2} is not reflected in the ``distance to goal''. 
On the other hand, once the parameters for \textbf{MLP} and \textbf{NPI}  are learned, a direct evaluation of the policy takes around $\SI{0.1}{\second}$  for all tasks, which is slightly faster than the evaluation of our GTN edges. 

\begin{figure}[t!]
\centering
\includegraphics[width=0.85\textwidth]{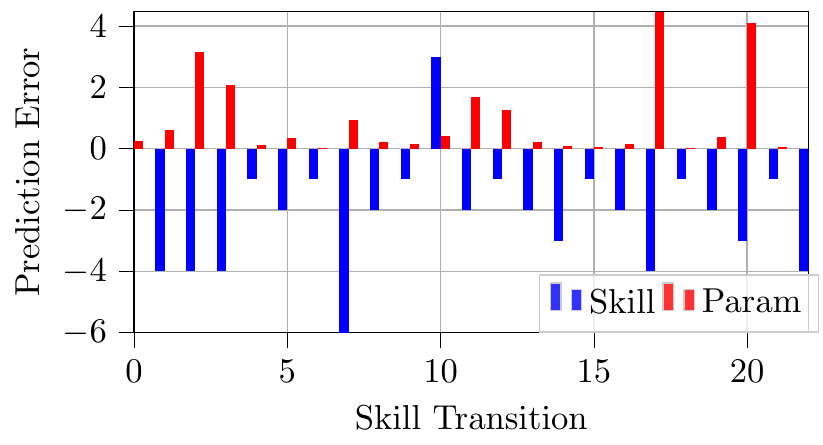}
\caption{
Prediction error during validation of method \textbf{NPI} for \textbf{Task-2}, regarding the choice of skill (in blue) and parameter (in red), indexed by the unique transitions in the plan. 
Negative error is the number of times when the skill choice is correct.
}
\label{fig:results-error}
\end{figure}

Last, the success rate during validation shows clearly the advantages of our method, especially for determining the skill parameters. 
As summarized in Fig.~\ref{fig:results-rate}, \textbf{GTN} has a consistent success rate of around $97\%$ for all tasks. 
When imposing a planning \emph{time limit} as $100$ times the solution time of the GTN planner, 
both \textbf{Hb-lfd} and \textbf{Hb-mp} perform relatively well for simpler tasks such \textbf{Task-1} and \textbf{Assembly}, while suffer from the high-dimensional search space for \textbf{Task-2} and \textbf{Task-3}.
On the other hand, the overall success rate for \textbf{NPI} and \textbf{MLP} is quite low (around $30\%$ for all tasks). 
As shown in Fig.~\ref{fig:results-train}, the testing accuracy for choosing the correct parameter is mainly the problem, as the RNN structure in \textbf{NPI} can reach $90\%$ accuracy when determining the next skill (compared to around $70\%$ for \textbf{MLP}). 
Furthermore, Fig.~\ref{fig:results-error} shows  the actual distribution of prediction error for method \textbf{NPI}. 
It is surprising to notice that: (I) The prediction for next skill is all correct except \emph{one} transition, i.e., transition $10$ from skill \texttt{STRT} to skill \texttt{Pi\_St}. 
As this transition is the beginning of a plan, the \textbf{NPI} network failed due to lack of past sequence and instead always chooses  the next skill \texttt{Pi\_Fl}.
(II) The prediction error for parameters is however high for \emph{all} transitions, especially when the same transition appears in the same plan multiple times but with different parameters. 
The underlying neural structure fails to capture such \emph{multi-modal} distribution.

\subsubsection{Limitations}\label{subsubsec:limitation}
In this section, we discuss some  limitations with the proposed approach, 
which are also parts of our ongoing and future work. 

\emph{Physical Interaction}.  The effect model in~\eqref{eq:skill-model} is used during the graph search to predict the resulting system state after executing a skill.  
Whereas it is efficient without constantly querying a simulator, it ignores the physical interaction during the execution. 
{Thus, we are currently investigating how forceful interactions can be added to the skill model~\cite{le2021learning}, 
and how a simulator can be requested on demand, e.g., to evaluate a proposed plan.}

\emph{Sub-task Modules}. It can be seen from the learned GTNs that often several skills can be grouped into sub-graphs that accomplish a sub-task of the complete task, e.g., the complete GTN of \textbf{Task-3} is composed by three sub-GTNs (one for each object), which is again very similar to the GTN learned for \textbf{Task-1}. 
It would be beneficial to recognize such sub-task modules in existing GTNs, which can be then composed into new GTNs during the learning process. 

\emph{{Generalization} and GTN Completion}. The learned GTN can only produce plans that are contained in the training set. 
{This can limit its ability to improve over existing solutions or generalize to new tasks, e.g., to different number and types of objects}.
The learned GTNs can be improved, e.g, via reinforcement learning 
by allowing deviations from the target policy and evaluating the outcome.


\section{Conclusions}\label{sec:conclusions}

A planning framework is proposed for coordinating manipulation skills that are learned from demonstrations. 
The learned policy is encoded as a GTN that encapsulates both the transition relations among skills and their underlying geometric constraints. 
It has shown a significant decrease in solution time and a much improved success rate, without any manual specification. 
Future work includes the structural composition of GTNs, 
{the online choice of pre-trained GTNs, and the generalization to new tasks}.


\section*{Acknowledgment}
The authors would like to thank Li-Yuan Hsu from University of Freiburg for his help on the setup of the simulation environment, during his internship at BCAI.

\bibliographystyle{IEEEtran}
\bibliography{contents/references}

\begin{IEEEbiography}[{\includegraphics[width=1in,height=1.25in,clip,keepaspectratio]{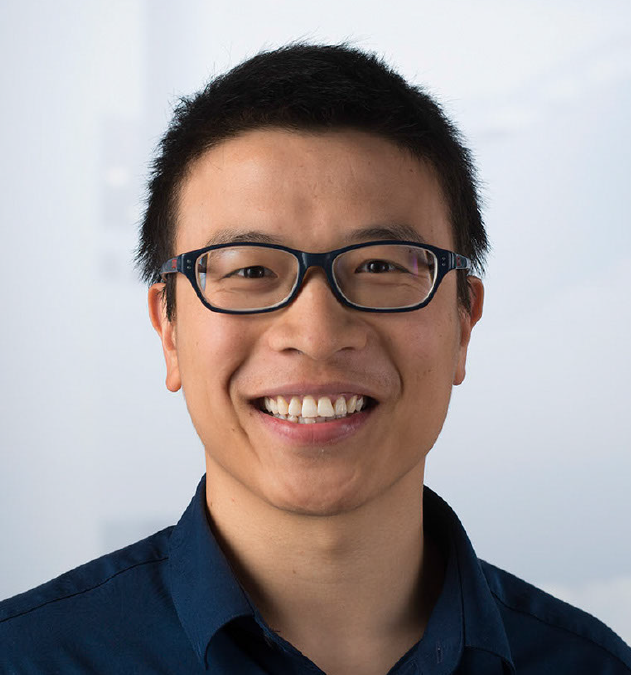}}]{Meng Guo}
received the M.Sc. degree (2011) in system, control, and robotics 
and the Ph.D. degree (2016) in electrical engineering from KTH Royal Institute of
Technology, Sweden.
He was a Postdoctoral Associate with the Department of Mechanical Engineering and Materials Science, Duke University, USA.
He is currently a senior research scientist on Reinforcement Learning and Planning at the Bosch Center for Artificial Intelligence (BCAI), Germany.
His main research interests include task and motion planning for robotic systems.
\end{IEEEbiography}

\begin{IEEEbiography}[{\includegraphics[width=1in,height=1.25in,clip,keepaspectratio]{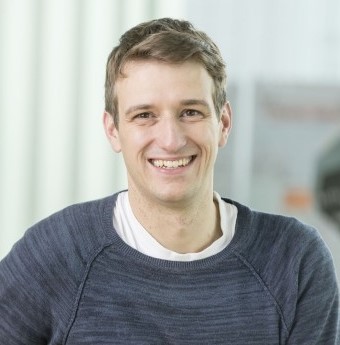}}]{Mathias B\"urger}
received his Diploma degree (2009) and his Ph.D. degree (2013) in Engineering Cybernetics from the University of Stuttgart. He received the 2014 European Ph.D. Award on Control for Complex and Heterogeneous Systems. He is currently head of a research group on Reinforcement Learning and Planning at the Bosch Center for Artificial Intelligence (BCAI). His research interest are within the intersection of Artificial Intelligence and Robotics. 
\end{IEEEbiography}


\end{document}